\documentclass{article}

\PassOptionsToPackage{numbers, compress}{natbib}

\usepackage[final]{neurips_2024}
\usepackage[utf8]{inputenc} 
\usepackage[T1]{fontenc}    
\usepackage{url}            
\usepackage{booktabs}       
\usepackage{amsfonts}       
\usepackage{nicefrac}       
\usepackage{microtype}      
\usepackage{xcolor}         
\usepackage{url}
\usepackage{graphicx}
\usepackage{svg}
\usepackage{animate}
\usepackage{listings}
\usepackage{algorithm}
\usepackage{algpseudocode}
\usepackage{enumitem}
\usepackage{silence}
\usepackage{makecell}
\usepackage{graphicx}  
\usepackage{wrapfig}   
\usepackage{caption}
\usepackage{subcaption}
\usepackage{float}
\usepackage{multirow}
\usepackage[accsupp]{axessibility}  

\newcommand{\figref}[1]{Fig.~\ref{#1}}
\newcommand{\tabref}[1]{Tab.~\ref{#1}}

\newcommand{\secref}[1]{Sec.~\ref{#1}}
\newcommand{\myPara}[1]{\noindent\textbf{#1}}

\usepackage[pagebackref,breaklinks,colorlinks,citecolor=gray]{hyperref}

\makeatletter
\renewcommand{\@noticestring}{\@neuripsordinal.}
\makeatother

\title{KVAE: Family of Tokenizers for Multimodal Generative Models}

\author{%
  Andrey Shutkin\textsuperscript{1}\\
  Kandinsky Lab\\
  \And
  Denis Parkhomenko\textsuperscript{1, 2, 3, 4}\\
  Kandinsky Lab\\
  \And
  Ivan Kirillov\textsuperscript{3}\\
  Kandinsky Lab\\
  \And
  Kirill Chernyshev\textsuperscript{1}\\
  Kandinsky Lab\\
  \And
  Kirill Malakhov\textsuperscript{2, 4}\\
  Kandinsky Lab\\
  \And
  Ilia Vasiliev\textsuperscript{2, 4}\\
  Kandinsky Lab\\
  \And
  Ilia Trushkin\textsuperscript{2}\\
  Kandinsky Lab\\
  \And
  Valeriya Kobenko\textsuperscript{2}\\
  Kandinsky Lab\\
  \And
  David Chikovani\textsuperscript{2, 4}\\
  Kandinsky Lab\\
  \And
  Alexander Ivanov\textsuperscript{3}\\
  Kandinsky Lab\\
  \And
  Azat Saginbaev\textsuperscript{3}\\
  Kandinsky Lab\\
  \And
  Egor Silvestrov\textsuperscript{3}\\
  Kandinsky Lab\\
  \And
  Ivan Mikheev\textsuperscript{1}\\
  Kandinsky Lab\\
  \And
  Konstantin Zakharov\textsuperscript{2}\\
  Kandinsky Lab\\
}

\newif \ifhq
\hqfalse

\begin{document}

\maketitle
\footnotetext[1]{worked on video tokenization}
\footnotetext[2]{worked on image tokenization}
\footnotetext[3]{worked on audio tokenization}
\footnotetext[4]{worked on diffusbility estimation}

\begin{abstract}

Latent diffusion modeling (LDM), a prominent paradigm, utilizes tokenizers to map input signal to compressed representation. This dependency positions tokenizer as an integral part of generation process itself, since it affects learning speed, quality of synthesized samples and lay foundation for later applications. This report presents series of KVAE tokenizers for audio, image and video, all designed for subsequent text-conditioned generation: KVAE-Audio, a continuous full-band 48 kHz tokenizer with a 50 Hz
latent of 64 channels; KVAE-3D -- two causal video tokenizers for 4x16x16 and 4x8x8 compression; KVAE-2D, an image model, compressing input by factor of 8 with 32 channels. We demonstrate that reconstruction (PSNR, LPIPS, PESQ, etc.) and generation results on objective (Frechet Distance, CLIP score, CLAP score, etc.) and subjective (side-by-side evaluation) metrics matches or surpasses frontier opensource tokenizers, such as VAEs from Wan-2.2, HunyuanVideo-1.5, FLUX.2, MovieGen, StableAudio and MMAudio. Considering difficulty of development, we share with community training details, model selection method and ablation on design choices. The code is publicly available at \url{https://github.com/kandinskylab/kvae} and \url{https://github.com/kandinskylab/kvae-audio}.
\end{abstract}

\section{Introduction}
\label{sec:intro}

In recent years development of generative models reached an unprecedented scale with evergrowing number of theoretical works and applications. Their fields of usage range from synthesizing molecules ~\citep{moleculegenie} and CT-scans ~\cite{ctscansmaisiv2} to Earth-weather prediction \citep{weatheraurora, weathergencast} and autonomous driving ~\cite{autonomousdrivingnvidia, autonomousdrivingwaymo}. Pinnacle of gradual shift toward adoption of this solutions lies in media content creation, meaning: audio ~\citep{audiogoogle, audiotangoflux}, images ~\citep{imagezimage, imagebitdance} and video ~\citep{wan21, ltx2}.

From technical standpoint, most of this works based on either autoregression or diffusion, including ones that combine both such as \citep{nextstep14b}. Consider text-conditioned diffusion-based video generation. This scenario differs from popular among researchers class-conditional or unconditional video generation (on UCF-101 \citep{ucf101} or SkyTimelapse \citep{skytimelapse}) in scale of compute, data and overall training difficulty, arising due to presence of additional modules, such as text encoder \citep{clip, flant5, qwen3}. High dimensinality of signal forces corresponding solutions (opensource as well as commercial) to generate videos in latent space following \citep{rombach2022ldm, blattmann2023stablevideo}. This space is formed via visual encoder and generated latents are transformed to plain videos through visual decoder, hence these pair encoder-decoder, named visual tokenizer, directly influence quality of generations, training dynamics and properties of generative model. While operating in entirely different space, audio tokenizer plays equally important role and share key characteristics with its visual counterpart.

Ability of tokenizer to form proper for diffusion latent space got recognized in ~\citep{improvingdiffusability} and was called \textit{diffusability}. Another important characteristic of tokenizer is \textit{compression ratio}, since it defines number of tokens generative model will operate upon. Due to quadratic complexity of attention, development of tokenizers with high compression ratio may be seen as form of optimization \citep{dcvideogen} apart from, for example, quantization \citep{quantturbodiffusion} or distillation \citep{distillationtmd}. So by targeting diffusability and high compression ratio simultaneously, one can speed up inference and training, which in turn reduces carbon emissions \citep{carbon1, carbon2}. Here we present series of novel models for multiple modalities under one common name KVAE, which matches or surpasses opensource alternatives on both generation and reconstruction, and also discuss problems, which arise in design process of efficient tokenizers for diffusion.

First sections of report (Sec. \ref{sec:related_works} - Sec. \ref{sec:analysis}) will be solely devoted to visual domain, which will lay down foundation for audio tokenizer in Sec. \ref{sec:audio}. It starts with discussion of prior art (Sec. \ref{sec:related_works}): evolution of video generation methods; overview of approaches for visual tokenization in image and video domains; estimates of tokenizer's diffusability. Then, we provide (Sec. \ref{sec:method}) details regarding proposed video models: choice of architecture, training strategy and optimization objectives. Content of next section focused on comparison of KVAE-3D models with opponnents on reconstruction (Sec. \ref{sec:rec_res}) and generation tasks (Sec. \ref{sec:img_gen} - Sec. \ref{sec:vid_gen}), including info on test datasets and overall procedure. Results section ends (Sec. \ref{sec:kvae2d_results}) with introduction and evaluation of KVAE-2D, an image tokenizer. Report continues with remarks on evaluation of diffusability in visual domain (Sec. \ref{sec:diffusability}) and ablatation in design choices (Sec. \ref{sec:ablation}), which offer some explanation for final configuration of KVAE video tokenizers. Content of last section (Sec. \ref{sec:audio}) reveals crucial details about novel KVAE audio tokenizer. Finally, after report's conclusion there is supplementary with visual examples and additional comments.

\section{Related Works}
\label{sec:related_works}

\subsection{Generative video models}

\textbf{Generative adversarial networks.}
Foundation for modern methods was laid in image domain and this principle holds for ~\citep{vondrick16}, where GAN ~\citep{thegan} was extended to unconditional video generation using spatio-temporal convolutions and background-foreground decomposition. Later works, such as Sync-DRAW ~\citep{syncdraw}, TGANs-C ~\citep{ganfromcaption}, IRC-GAN ~\citep{ircgan} set more ambitious goal of generating with text condition, where the latter was encoded by recurrent networks. And results of DVD-GAN ~\citep{clark19} only confirmed that significant amount of training data is needed for advesarial setting in order to generate consistent videos. Computational burden of video modelling motivated MoCoGAN-HD ~\citep{ganhighressynth} to seek motion in latent space of frozen image generator, which speed up training and enable native inference for higher resolutions frames (up to 1024x1024 in the paper). Both StyleGAN-V ~\citep{styleganv} and ~\citep{ganlongvids} not only improved fidelity of generations, but also targeted longer videos, yet they've retreated to simpler tasks of unconditional generation, which raises scalability questions.

\textbf{Autoregression.}
First attempts, such as VPN ~\citep{vpn} adapt PixelCNN ~\citep{pixelcnn} and model each R/G/B-color pixel by pixel. In order to make this viable, VideoGPT ~\citep{videogpt} proposed modelling with VQ-VAE ~\citep{vqvae}, utilizing its rich codebook of visual representations, and substituted recurrent neural networks with Transformer ~\citep{thetransformer} as main computational block, enhancing effective context size. By scaling data and model size CogVideo ~\citep{cogvideo} was able to generate short 480x480 videos from text description. While this works targeted only video domain, VideoPoet ~\citep{videopoet} and Emu3 ~\citep{emu3} pursued the goal of modalities unification under LLM-based pipeline. Generation quality of next-token prediction models continue to grow and spark interest from large-scale projects, such as GrokImagine ~\citep{grokimagine} where it successfully applied.

\textbf{Diffusion.}
VDM ~\citep{vdm} was among first to view text-to-video from conditional diffusion perspective. It was followed by Make-a-video ~\citep{makeavideo}, where videos were treated in unsupervised manner, leveraging existing text-visual alignment of T2I models. Utilizing cascaded diffusion, Imagen Video~\citep{imagenvideo} set new generation standard by synthesizing 5 seconds of 24FPS video in HD resolution. While delivering state-of-the-art results, it was still pixel-based, but LDM ~\citep{ldm} was eventually extended to video in Video LDM~\citep{alignyourlatents} and StableVideo~\citep{stablevideo}. This latter was able to train text-to-video LDM from scratch by dataset scaling, thorough filtering and tree-stage curriculum. Shortly after, unleashing diffusion potential, Sora ~\citep{soraopenai} video generation model was anounced. Since then multitude solutions have emerged, proprietary (Kling~\citep{klingai}, Runway~\citep{runway_gen4}, Veo~\citep{google_veo31}) as well as opensource (Kandinsky~\citep{kandinsky5}, HunyuanVideo~\citep{hunyuanvideo}, Wan~\citep{wan21}, LTX~\citep{ltx2}). While high-quality video generation is still challenging (e.g. duration), some of this products incorporate synced audio generation~\citep{ltx2} or enable precise editing~\citep{moviegen}, pushing boundaries of generative media.

\subsection{Visual tokenizers for LDM}

Modeling in latent space of VAE ~\citep{vaekingma} was proposed in original LDM work ~\citep{ldm} for text-to-image task.
This two stage decomposition creates dependency on properties and quality of tokenizer, which in turn results in research directions seeking useful modifications, such as: compression ratio ~\citep{dcae, dcae15}, incorporation of visual foundation model ~\citep{rae, uae}, diffusion-based decoders ~\citep{flowtomode, ssdd}.
Meanwhile, video models were challenged with temporal compression: from Stable Video ~\citep{blattmann2023stablevideo} completely discarding it to experimental CV-VAE ~\citep{cvvae}, inheriting latent space from pretrained image model, and Reducio ~\citep{reducio}, performing compression with condition on keyframes akin to traditional video codecs.
Yet mainstream works, among which are CogVideoX ~\citep{cogvideox}, Cosmos ~\citep{cosmos} and WF-VAE ~\citep{wfvae}, are causal tokenizers ~\citep{langmodelbeatsdiffusion} trained simultaneously on images and videos, capable to compress number of frames in 4 times at least.
Progress in the field highlighted by the fact that every new generation of popular opensource solutions (Wan~\citep{wan22}, Hunyuan~\citep{hunyuan15}) includes updated tokenizer with improved metrics, featuring changes in compression ratio and architecture.

\subsection{Generative performance prediction}
  \label{sec:gen_perf_pred}
  A practical bottleneck in LDM development is selecting a visual tokenizer before incurring the
  cost of training a full downstream diffusion model.  Reconstruction metrics are available at
  this stage, but they do not reliably predict downstream generation quality.  Yao et
  al.~\citep{yao2025reconstruction} describe this conflict as the reconstruction--generation
  dilemma: under fixed downstream training compute, increasing the per-token feature dimension can
  improve reconstruction FID while degrading generative FID.  Their VA-VAE aligns latents with
  vision foundation model features and expands the observed reconstruction--generation Pareto
  frontier.  These findings motivate the study of latent-space properties beyond reconstruction
  fidelity.\par
  \textit{Diffusability} characterizes how amenable a latent representation is to diffusion
  modeling, rather than how accurately its decoder reconstructs the
  input~\citep{improvingdiffusability}.  Diffusability should not be conflated with generative
  performance itself: it describes a structural property of the latent space, while frequency and
  correlation structure are candidate diagnostics for this property rather than its definition.
  Skorokhodov et al.\ report unusually strong high-frequency components in modern autoencoder
  latents, especially in high-dimensional bottlenecks, and hypothesize that these components
  interfere with the coarse-to-fine bias of diffusion synthesis.  Their scale-equivariance
  regularizer aligns latent- and RGB-space frequency behavior and improves image and video
  generation after short fine-tuning.  Reconstruction fidelity, diffusability, and generative
  performance should therefore be treated as three distinct axes, and how well the first two
  predict the third remains an open question.\par
  Subsequent work develops more specific spectral hypotheses along two separate axes.  For video
  VAEs, SSVAE identifies a spatio-temporal spectrum biased toward low frequencies and, separately,
  a channel eigenspectrum concentrated in a small number of modes~\citep{liu2025ssvae}.  These
  are distinct properties---one concerns spatio-temporal frequency content, the other the
  distribution of variance across channels---and SSVAE's Local Correlation Regularization and
  Latent Masked Reconstruction objectives target both.  The authors report a $3\times$
  acceleration of text-to-video convergence and a $10\%$ improvement in video reward over the
  evaluated open-source baselines.  Spectrum Matching~\citep{ning2026spectrummatching} instead
  characterizes the target spectrum as flattened but still power-law decaying, rather than simply
  concentrated at low frequencies, and adds a capacity constraint: this shape must be balanced
  against information preservation under a fixed bottleneck, since excessive smoothing removes
  detail whereas excessive whitening can impede denoising.  Encoding Spectrum Matching constrains
  the latent PSD toward a flattened power-law shape, whereas Decoding Spectrum Matching preserves
  frequency-to-frequency semantic correspondence through the decoder.\par
  REPA-E provides complementary end-to-end evidence using two contrasting
  VAEs~\citep{leng2025repae}.  REPA-E end-to-end tuning makes the noisy SD-VAE latents smoother
  and the over-smoothed IN-VAE latents more detailed.  By contrast, directly backpropagating the
  diffusion loss into the VAE encourages simpler latent structures that are easier to denoise but
  degrade generation quality.  Together, these studies suggest a working hypothesis: a generative
  latent may need enough spectral bias to support diffusion modeling while retaining enough
  information capacity for faithful reconstruction and synthesis.  Diffusability and
  reconstruction fidelity should therefore be treated as jointly constrained requirements rather
  than a simple trade-off, although the precise mechanism linking spectral bias and information
  capacity to generation quality is not yet established.\par
  These findings motivate testing inexpensive frozen-tokenizer diagnostics that could plausibly
  reflect this spectral-bias--information-capacity balance.  In Sec.~\ref{sec:diffusability}, we
  evaluate one such candidate, a spatial correlation-decay statistic (CDS), and examine its
  association with downstream generation quality as an initial candidate-screening criterion.\par

\section{Method}
\label{sec:method}
\subsection{Definitions}

Define input RGB video as $x \in R^{3\times T\times H\times W}$, then visual tokenizer is a pair of encoder $\mathcal{E}$ compressing original signal and decoder $\mathcal{D}$ reconstructing it. Specifically:
\begin{gather*}
\mathcal{E}: R^{3\times T\times H\times W}\rightarrow R^{c\times t\times h\times w}\\
\\
\mathcal{D}: R^{c\times t\times h\times w} \rightarrow  R^{3\times T\times H\times W}\\
\\
t = \frac{T-1}{f_t} +1 ;\quad h = \frac{H}{f_s} ;\quad w = \frac{W}{f_s}
\end{gather*}

Factors $f_t$, $f_s$ and number of hidden channels $c$ are fixed by layers configuration and may serve as charachteristic of tokenizer. Here and later, when referencing some tokenizer its compression parameters will be put together in short string of form $f_t \times f_s \times f_s \ (c\  \text{channels})$. Namely, this work presents two video models: $\text{KVAE--}4\times 8\times 8\ (16\ \text{channels})$ and $\text{KVAE--}4\times 16\times 16\ (64\ \text{channels})$.

\subsection{Architecture}

While video tokenizer's architecture often appear similar to the image ones, there is also a significant differences because of the additional temporal axis. First, while input and output convolutions may remain 2D ~\citep{h3ae}, deeper layers are either plain Conv3D ~\citep{cogvideox} or decomposed on temporal and spatial components ~\citep{cosmos}. Second, downsample and upsample are also operate on this new axis, raising issues of compression order and block's design. Third, targeting video generation progressive training schedule and final use-cases, such as Image-to-Video, tokenizer should process images and videos, which results in causal setup, usually implemented with assymetric temporal padding. Fourth, attention choice affects training and inference, with options such as full attention ~\citep{hunyuanvideo}, spatio-temporal ~\citep{cosmos} and spatial-only ~\citep{wan22}. Finally, for long sequences and large resolutions tiling (spatial or temporal)  almost certainly will be used.

Insipired by CogVideoX~\citep{cogvideox}, proposed KVAE models are trained to compress long sequences of frames. For that reason, their design is attention-free with Conv3D as main computational block. This choice leads to efficient training and ability to perceive arbitrarily long sequences on inference via caching mechanism. As for normalization, both models built upon spatial $\text{RMSNorm}$, which allows segment size adjustment on inference. Motivated by ~\citep{dcae, cosmos} downsample and upsample blocks perform temporal and spatial operations sequentially. Important to note that while Wan-2.2 ~\citep{wan22} applies patching as form of compression on earlier layers, KVAE-4x16x16 leaves it for downsample blocks.

Among visual tokenizers, assymetry in number of parameters between encoder and decoder is common. For KVAE-4x8x8 it takes moderate form -- with more layers decoder is approximately $1.3\times$ bigger than encoder. But KVAE-4x16x16 takes step further by making decoder layers wider and reducing channels in encoder layers. It results in parameters ratio of $5.3\times$, which is explained by necessity of higher generation capabilities of decoder.

\subsection{Training methodology}

Since production-grade solutions target generation of 10 seconds videos with 24FPS in FullHD resolution, tokenizer should perceive such inputs reliably. For that reason on training we adapt sequence length scaling approach from CogVideoX~\citep{cogvideox}. But instead of starting with 17 frames, it begins with 65 frames, which throughout our experiments led to faster convergence. Moreover, we found that limiting number of frames during training by 129 is enough and models generalize well to longer sequences (up to 400 on test), allowing larger batch size than with original limit of 161 frames. While causal architecture already allows to perceive single frame, during training video batches were interleaved with images with batch sampling probability of 0.3.

Training for both models is divided into 4 stages defined by their objectives. On first stage sum of 3 components is considered:
\[
L^I = \mathcal{L}_1(x, \mathcal{D}(\mathcal{E}(x)_s)) + w_{\text{perc}}\cdot\text{LPIPS}(x, \mathcal{D}(\mathcal{E}(x)_s)) + w_{\text{KL}}\cdot D_{KL}(\mathcal{E}(x), \mathcal{N}(0, I))
\]
where $\mathcal{L}_1$ is a Mean Average Error, $\text{LPIPS}$ ~\citep{lpips} is well-known perceptual loss, $D_{KL}$ is Kullback-Leibler divergence and $\mathcal{E}(x)_s$ denotes sample from normal distribution with parameters $\mathcal{E}(x)$. Second stage introduces frame-based GAN ~\citep{vqgan} with few thousand warm-up steps for stability:
\[
L^{II} = L^I + w_{adv}\cdot \text{GAN}(\mathcal{D}(\mathcal{E}(x)_s), x)
\]
Third stage adapt EQ-loss ~\citep{eqvae} on top of $L^{II}$ with latent downsampling, rotation and last stage is decoder finetuning.
Objectives are optimized by Adam~\citep{kigmaadam} with learning rates being fixed for every stage.

Training dataset for tokenizer comprises of 10 million images and 2 million videos, which all were chosen from pretrain dataset of generative model. Accounting for only a few percentages of total data, its content already filtered by resolution, motion and content. Preserving characteristics of original signal, we limit preprocessing transformations to flip, rotation, resize and crop with resolution varying across the stages from 256 to 512.

\section{Results}
\subsection{Reconstruction}
\label{sec:rec_res}

Main purpose of tokenizers is to provide latent space for generative models.
And, as was shown in series of works regarding image domain~\citep{recvsgen, rae}, excellent reconstructions does not necessarily align with high generation results.
Yet, since proposed models were trained under reconstruction objectives, corresponding metrics would indicate their ability to build visual representations and absence of artifacts after decoding.

\begin{table}[H] 
    \centering
    \caption{Reconstruction results for tokenizers on MCL-JCV 720p, grouped by compression factor}
    \label{tab:rectable}
    \begin{tabular}{lccccc}
        \toprule
        \textbf{model} & \textbf{factor} & \textbf{channels} & \textbf{PSNR} & \textbf{SSIM} & \textbf{LPIPS} \\
        \midrule
        HunyuanVideo-1.0 & \multirow{3}{*}{\centering 4x8x8} & \multirow{3}{*}{\centering 16} & 34.3 & 0.90 & 0.047 \\
        Wan-2.1 & & & 34.3 & 0.89 & \textbf{0.044} \\
        KVAE-2.0 & & & \textbf{36.0} & \textbf{0.92} & 0.047 \\
        \midrule
        HunyuanVideo-1.5 & \multirow{3}{*}{\centering 4×16×16} & 32 & 34.4 & 0.89 & 0.073 \\
        Wan-2.2 & & 48 & 34.2 & 0.89 & \textbf{0.037} \\
        KVAE-2.0 & & 64 & \textbf{35.2} & \textbf{0.91} & 0.058 \\
        \bottomrule
    \end{tabular}
\end{table}

We compare against leading opensource methods, such as HunyuanVideo and Wan, on MCL-JCV\citep{mcljcv}, chosen as primary source of test video sequences, because of resolution and length in frames.
Sequences were trimmed until total amount of frames can be expressed in form $8\times k + 1$, where $k \in \mathbb{N}$.
Since HunyaunVideo tokenizers using full attention, on long sequences tiling was applied to their inputs and latents. As for Wan and KVAE -- no modifications to the original forward procedure were made. And as can be seen in Table \ref{tab:rectable}, proposed models gain additional PSNR (on average higher than 1 dB) with negligible LPIPS loss. 

\begin{figure}[htbp]
    \centering
    \includegraphics[width=0.9\linewidth]{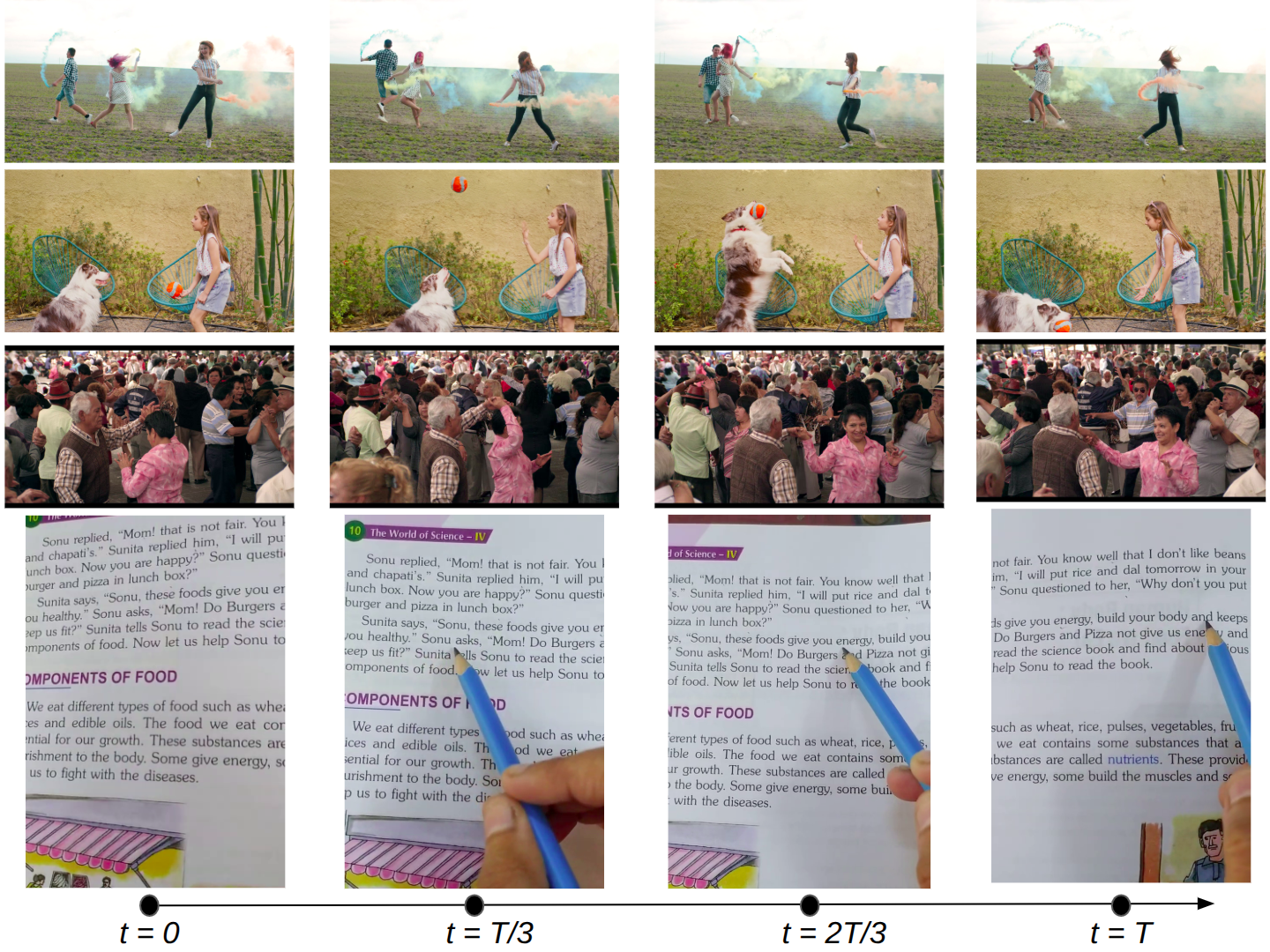}
    \caption{Frames from internal video dataset used for intermediate evaluations}
    \label{fig:dataset}
\end{figure}

Despite the proposals for tokenizer's dataset, such as TokenBench in~\citep{cosmos}, there is still no common test procedure.
So by composing few open datasets, such as AOM-CTC ~\citep{aom_av2_ctc_v5}, BVI-DVC ~\citep{bvidvc} and others, we harvested a few hunderds videos varying by their content, resolution and motion (see Fig. \ref{fig:dataset}).
Being divided by categories, this dataset allowed extensive evaluation during development of the models.

\subsection{Image generation}
\label{sec:img_gen}

Following procedure described in Kandinsky-5 ~\citep{kandinsky5} training of video generation model starts from text-to-image stage on low resolution.
This model is built upon CrossDiT architecure, utilizing Qwen-2.5VL~\citep{qwen25vl} for text embeddings and minimizes flow matching loss.
All image and video experiments were conducted on 2B generation model. 
Because visual encoders considered here have different compression factors, it is possible to alter patch size, balancing amount of input visual tokens.
Specifically, for KVAE-4x8x8 patch size of 2x2 is set, while to all 4x16x16 models patch 1x1 is applied.
With same dataset (of few hundred millions images), captions and hyperparameters, the final results are expected to depend only on visual tokenizers.

\begin{figure}[htbp]
\centering
\includegraphics[width=0.95\linewidth]{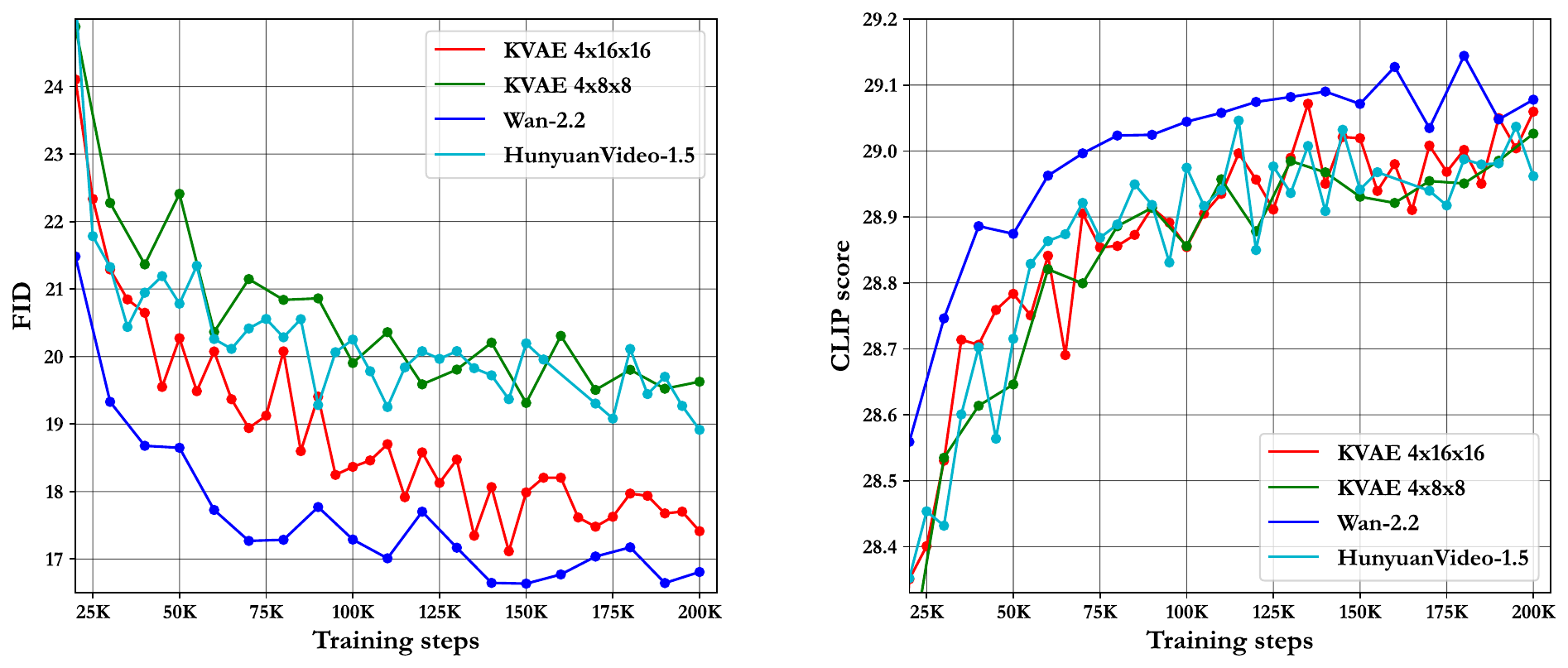}
\caption{Learning curves of text-to-image generation models on 384x256 resolution}
\label{fig:image_gen_dynamic}
\end{figure}

Generations are evaluated by CLIP~\citep{clip} and FID~\citep{fid} every 10K steps of optimizer, see Fig. \ref{fig:image_gen_dynamic} for learning dynamic.
From it a conclusion can be made that tokenizers with higher compression factors bring faster convergence.
Increasing number of channels in KVAE-4x16x16 up to 64 provided benefits, such as surpassing FID results of HunyuanVideo-1.5 with 32 channels, which raises an open question on relation between compression ratio and generative performance.

In addition to CLIP and FID, for final checkpoints side-by-side evaluation is performed on rich set of prompts, ranging across styles (photorealism, futuristic, cartoon, etc.), scenery (interior, outdoors and more), subjects (their number, appearance).
For each pair, generated from same prompt, annotator mark preferable image independently for three categories: prompt following, visual quality and semantic quality. If it is impossible to decide, then there is draw options (\textit{both good} or \textit{both bad}) and exact formula for win rate is:

\[
\text{WinRate}(A) = \frac{\text{win}_A + 0.5 \cdot \text{both good}}{\text{win}_A + \text{win}_B + \text{both good}}
\]

\begin{figure}[htbp]
\centering
\begin{subfigure}[t]{0.49\textwidth}
\centering
\includegraphics[width=\linewidth]{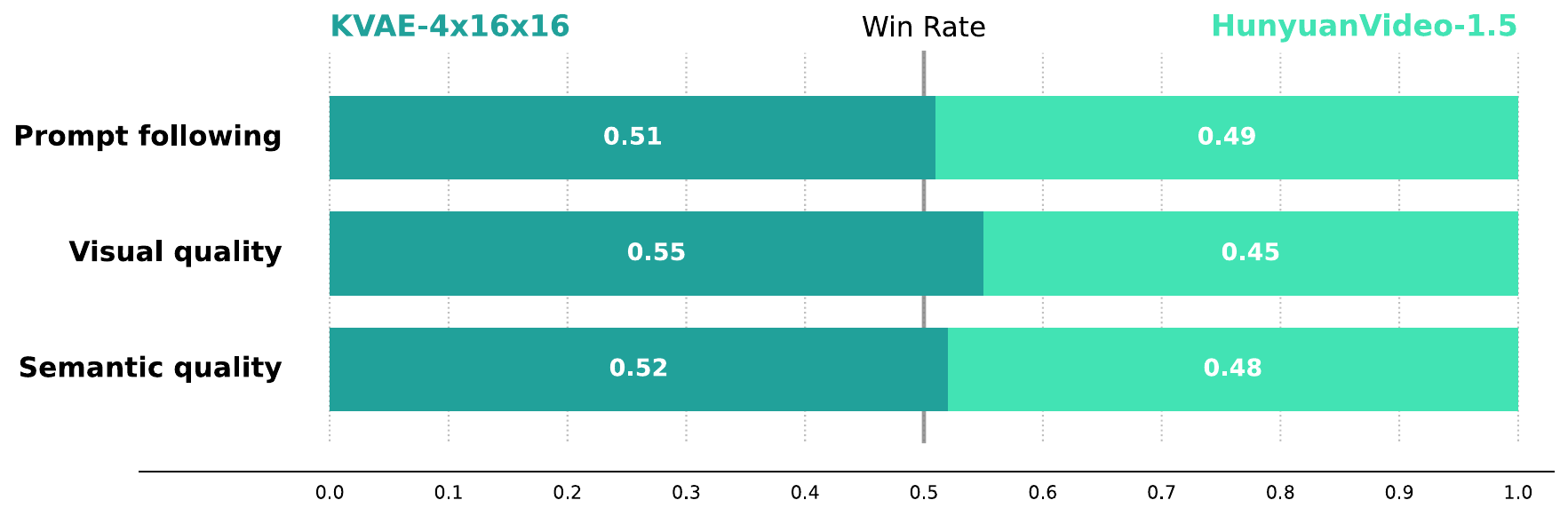}
\caption{Comparison with HunyuanVideo-1.5}
\end{subfigure}
\hfill
\begin{subfigure}[t]{0.49\textwidth}
\centering
\includegraphics[width=\linewidth]{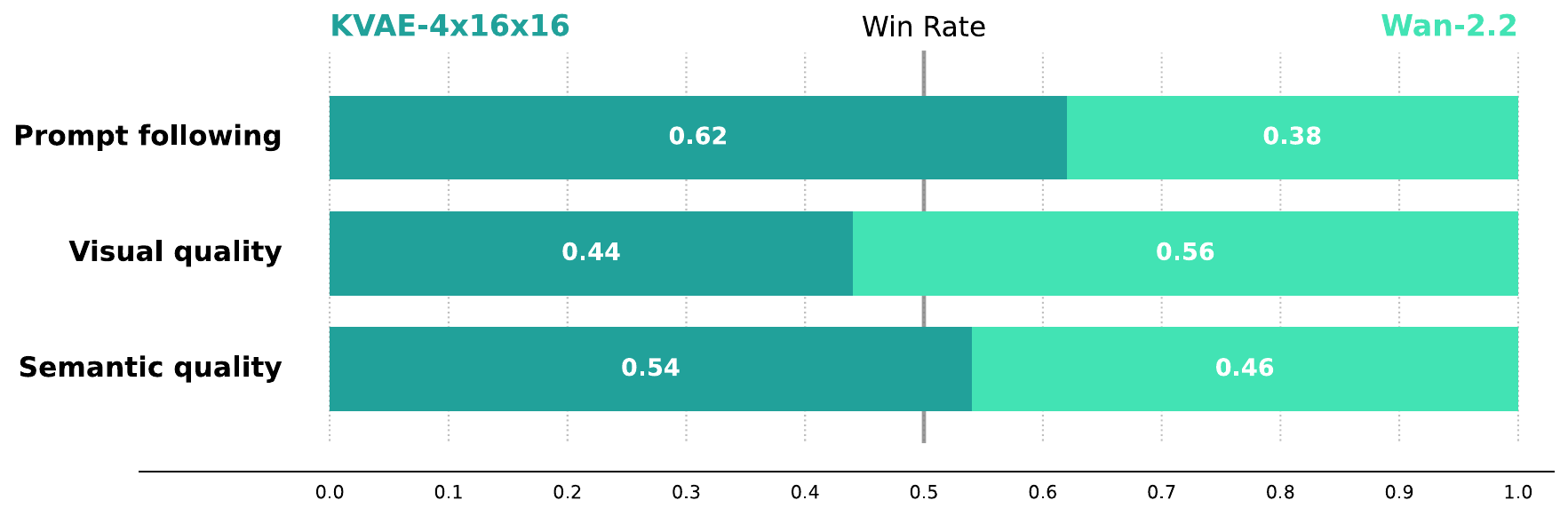}
\caption{Comparison with Wan-2.2}
\end{subfigure}
\caption{Results of side-by-side evaluation of image generations for KVAE-4x16x16}
\label{fig:sbs_416}
\end{figure}

\begin{figure}[htbp]
\centering
\begin{subfigure}[t]{0.49\textwidth}
\centering
\includegraphics[width=\linewidth]{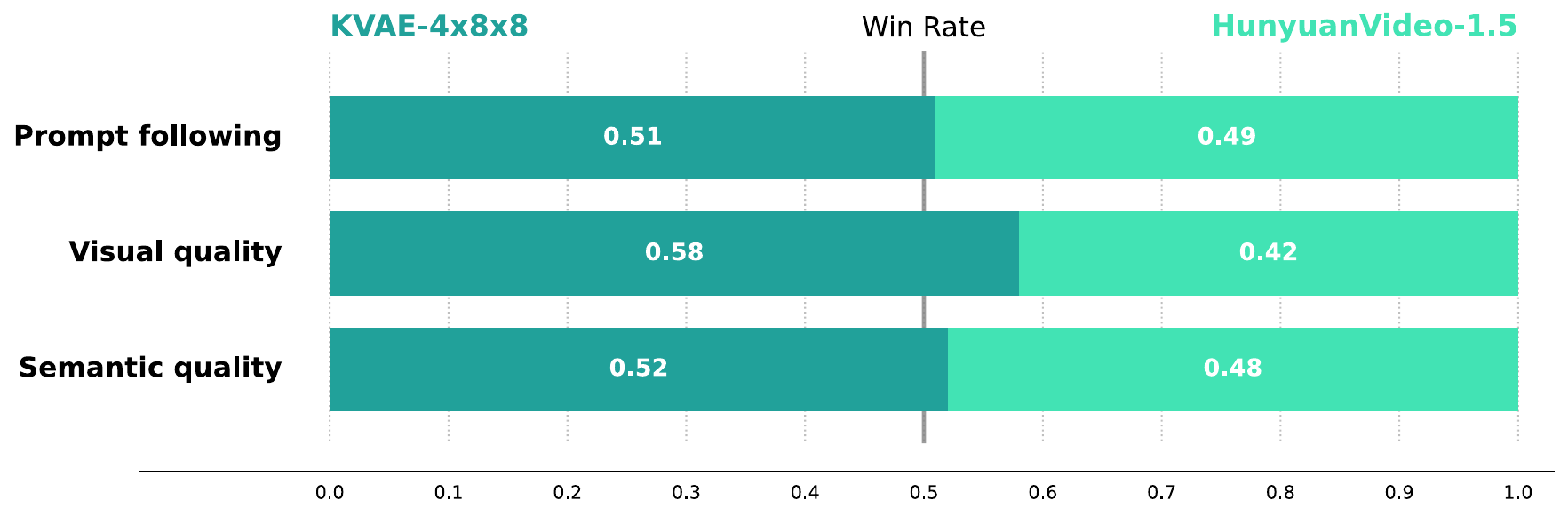}
\caption{Comparison with HunyuanVideo-1.5}
\end{subfigure}
\hfill
\begin{subfigure}[t]{0.49\textwidth}
\centering
\includegraphics[width=\linewidth]{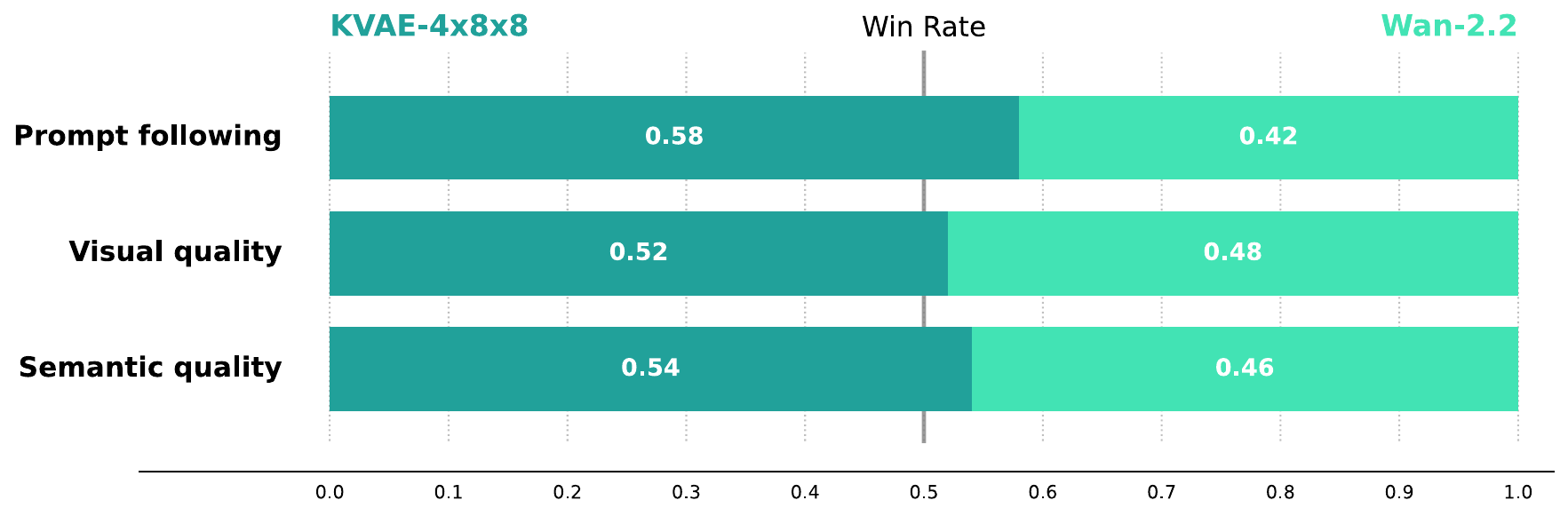}
\caption{Comparison with Wan-2.2}
\end{subfigure}
\caption{Results of side-by-side evaluation of image generations for KVAE-4x8x8}
\label{fig:sbs_488}
\end{figure}

Surprisingly, for KVAE-4x8x8 (see Fig. \ref{fig:sbs_488}) objective metrics contradict with side-by-side evaluation, where it holds leading positions. At the same time, KVAE-4x16x16 chosen in total accordance with its metrics (see Fig. \ref{fig:sbs_416}): gives up visual appearance over prompt following in comparison to Wan-2.2 and consistenly preferred over HunyuanVideo-1.5 on all categories.

\subsection{Video generation}
\label{sec:vid_gen}

\begin{figure}[H]
\centering
\includegraphics[width=0.75\linewidth]{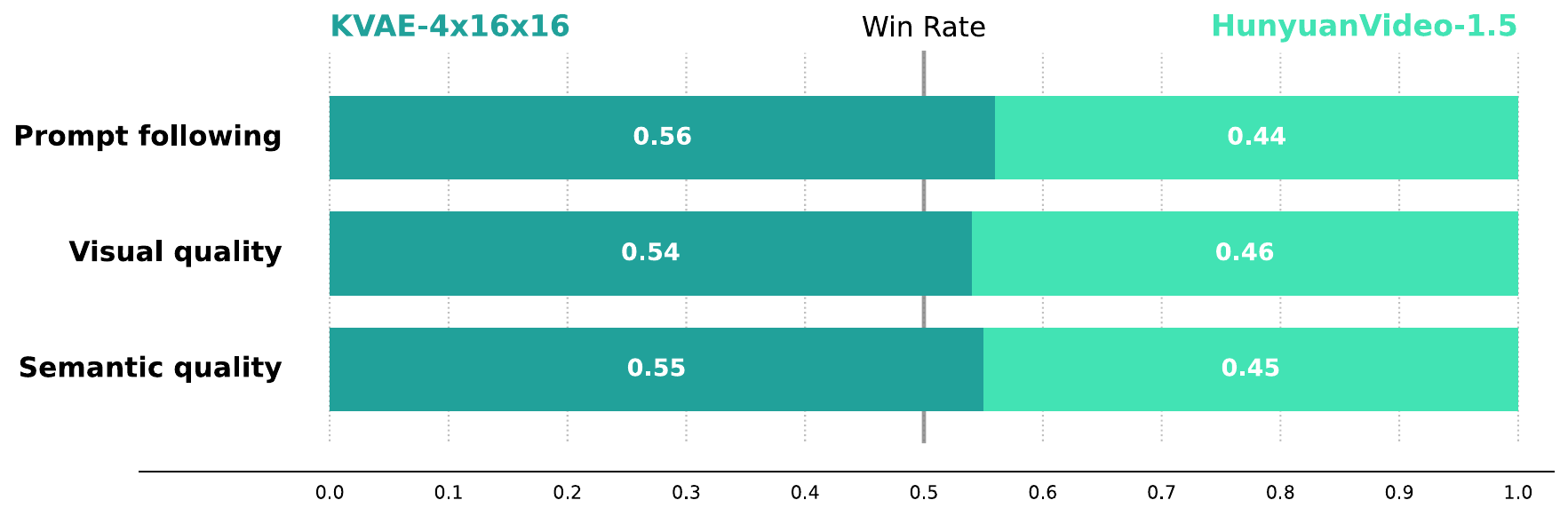}
\caption{Side-by-side evaluation of 121-frame 768x512 video generations for KVAE-4x16x16 and HunyuanVideo-1.5}
\label{fig:video_results}
\end{figure}

\begin{figure}[H]
\centering
\includegraphics[width=0.95\linewidth]{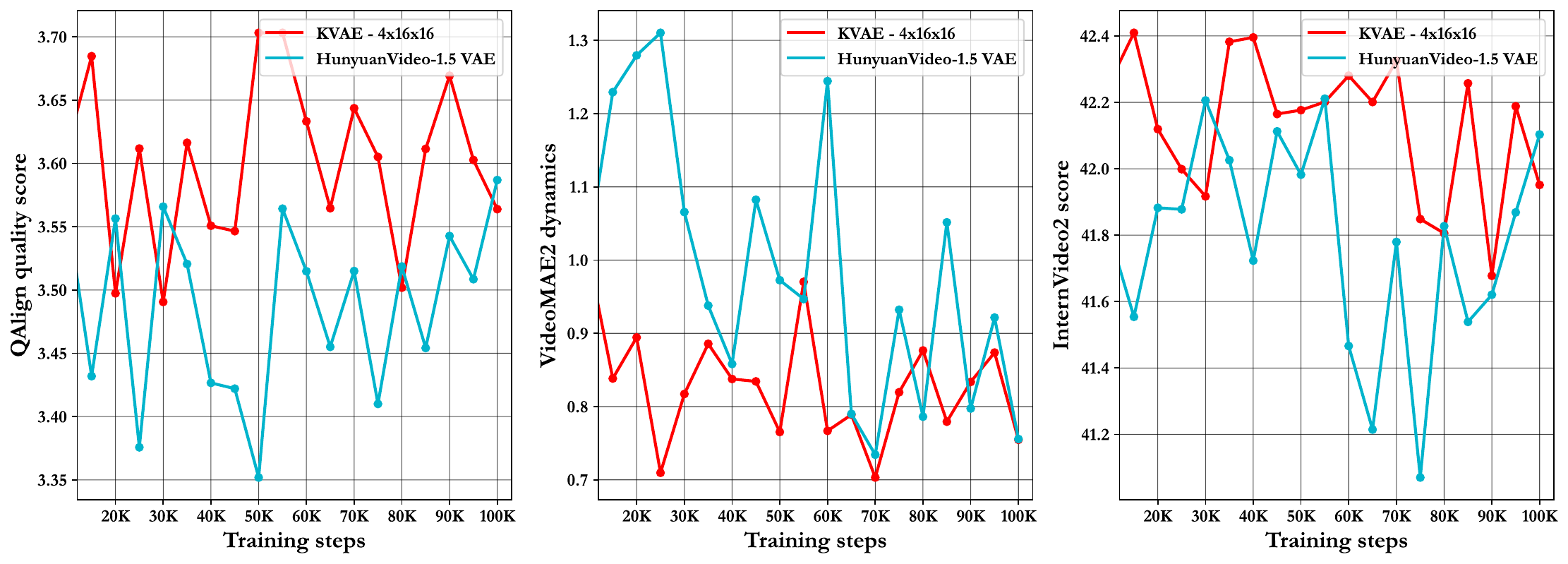}
\caption{Learning curves of text-to-video generation models on 121-frame 384x256 resolution. With higher values of QAlign and InterVideo2 scores, training with KVAE-4x16x16 is also characterized by less variance across this three metrics.}
\label{fig:video_generation}
\end{figure}

Image checkpoint serves as starting point for video generation training, which progressively increase resolution across stages.
To monitor intermediate quality of generated videos we evaluate such metrics as text-to-video correspondence through InternVideo2~\citep{internvideo2}, overall visual appearance with QAlign~\citep{qalign} and dynamics estimation based on VideoMAE2 features~\citep{videomae2, dkoposovmotion}.
Comparison of HunyaunVideo-1.5 VAE and KVAE-4x16x16 during validation on resolution 384x256 are provided on Fig. \ref{fig:video_generation} and results of side-by-side evaluation on resolution 768x512 are present in Fig. \ref{fig:video_results}.

For text prompts we utilize MovieGen Benchmark~\citep{moviegen}, where each query is enriched with object details and motion description, allowing to gather a set of diverse videos from almost thousand prompts.
Results of this experiments allow to conclude that under Kandinsky-5 training pipeline, featuring large amount of filtered data, several stages and valid multimodal architecture, KVAE-4x16x16 form better latent space than HunyuanVideo-1.5 VAE, which eventually led to faster convergence and better generations.

 \subsection{KVAE-2D-2.0 results}
  \label{sec:kvae2d_results}
  To complement the video-tokenizer results, we evaluate \textbf{KVAE-2D-2.0}, the final image
  tokenizer used in the image-domain diffusability analysis of Sec.~\ref{sec:diffusability}.
  Architecturally, it is a four-level convolutional residual autoencoder with base width 128 and
  channel multipliers $(1,2,4,8)$. It maps images to a 32-channel Gaussian latent with 8x8 spatial
  compression.\par
  We first evaluate tokenizer reconstruction on OmniDoc-TokenBench, a 3042-image benchmark of
  text-rich documents~\citep{zhang2026qwenimagevae20,alibaba2026omnidoc}.
  Table~\ref{tab:omnidoc_kvae2d_flux} reports KVAE-2D-2.0 surpassing published
  FLUX.1-dev and FLUX.2-dev baselines on PSNR, SSIM and NED.
  \begin{table}[H]
  \centering
  \scriptsize
  \caption{OmniDoc-TokenBench reconstruction results. FLUX.1-dev and FLUX.2-dev are published third-party baselines.}
  \label{tab:omnidoc_kvae2d_flux}
  \resizebox{\linewidth}{!}{%
  \begin{tabular}{lccc}
  \toprule
  Metric &
  KVAE-2D-2.0 (ours, 8x8, 32 channels) &
  FLUX.1-dev (published, 8x8, 16 channels) &
  FLUX.2-dev (published, 8x8, 32 channels) \\
  \midrule
  PSNR $\uparrow$  & \textbf{28.05}  & 26.24            & 27.72 \\
  SSIM $\uparrow$  & \textbf{0.957}  & 0.9364           & 0.9544 \\
  LPIPS $\downarrow$ & 0.028         & 0.0247           & \textbf{0.0216} \\
  FID $\downarrow$   & 1.74          & \textbf{0.5543}  & 0.73 \\
  NED $\uparrow$     & \textbf{0.976} & 0.9546          & 0.9535 \\
  \bottomrule
  \end{tabular}%
  }
  \end{table}

    Motivated by the strong public T2I performance of FLUX.2 reported in
  DiffusionBench~\citep{diffusionbench2026}, we use its native VAE as the reference tokenizer. We
  compare generations from two downstream 2B T2I DiT checkpoints after 200K optimization steps:
  one stack uses the FLUX.2 VAE, and the other uses the 400K-step KVAE-2D-2.0 checkpoint.
  The point estimates favor the KVAE-2D-2.0 stack in all three criteria, with the largest
  difference in semantic quality (5.3 percentage points).\par

  \begin{figure}[H]
  \centering
  \includegraphics[width=\linewidth]{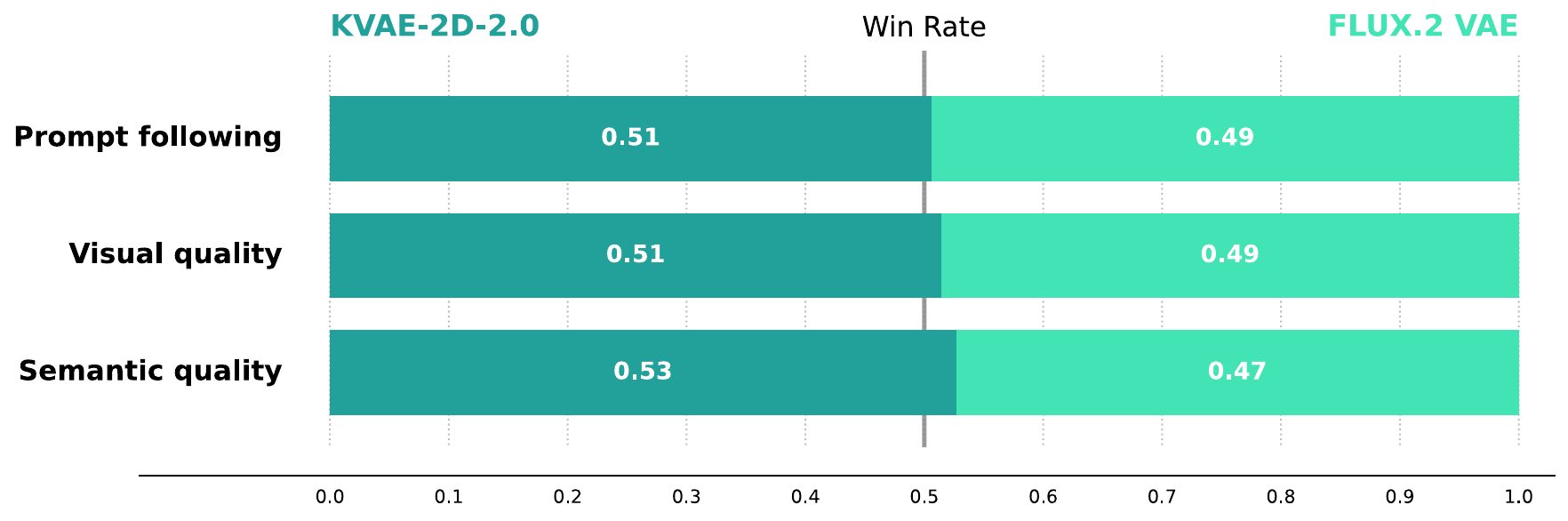}
  \caption{Side-by-side soft win rates for the two downstream 2B T2I DiT stacks at 200K
  optimization steps.}
  \label{fig:sbs_kvae2d_flux2}
  \end{figure}

\section{Analysis}
\label{sec:analysis}

During development phase a number of design choices were filtered out, mainly based on quality of generation models trained on top of corresponding tokenizers. This process eventually led to final weights of KVAE-2D-2.0, KVAE-4x16x16 and KVAE-4x8x8. Below we provide details regarding diffusability estimation, as a way to select best tokenizers prior to generation model training, and comparison with alternative versions of KVAE, confirming superiority of proposed models.

 \subsection{Diffusability}
  \label{sec:diffusability}
  \myPara{Screening setup.}
  Building on Sec.~\ref{sec:gen_perf_pred}, we examine associations between inexpensive
  measurements of a frozen tokenizer and the visual quality of generations produced by a
  downstream DiT.  We evaluated spatial self-similarity statistics, semantic probes, spectral
  and PCA-based diagnostics, drawing on iREPA~\citep{singh2025irepa}, SSVAE~\citep{liu2025ssvae},
  and Spectrum Matching~\citep{ning2026spectrummatching}.  This section focuses on Correlation
  Decay Slope (CDS), the primary spatial statistic retained for model selection, and reports its
  behavior and supporting evidence.\par
  \myPara{Subjective target.}
  The target is the visual-quality criterion from human side-by-side evaluations.  We aggregate
  pairwise outcomes with a Bradley--Terry model~\citep{bradley1952rank}, treating \emph{both
  good} and \emph{both bad} as one half-win for each system.  This differs from the descriptive
  win rate in Sec.~\ref{sec:img_gen}-~\ref{sec:kvae2d_results}, which assigns one half-win to \emph{both good} but
  excludes \emph{both bad}.  The analysis spans 14 tokenizer configurations; the Bradley--Terry
  refit incorporates the KVAE-2D-2.0 versus FLUX.2 comparison reported in
  Sec.~\ref{sec:kvae2d_results}.  The resulting target captures subjective downstream visual
  quality rather than reconstruction fidelity.\par
  \myPara{Correlation Decay Slope.}
  Let $z_{x,p}\in\mathbb{R}^{C}$ denote the latent feature vector for image $x$ at spatial
  position $p$. Using cosine similarity, we define the spatial correlogram and CDS
  as:\\
  {\setlength{\abovedisplayskip}{4pt}\setlength{\belowdisplayskip}{4pt}%
  \setlength{\abovedisplayshortskip}{0pt}\setlength{\belowdisplayshortskip}{4pt}%
  \[
  g(\delta)=\mathbb{E}_{x,p,q}\Big[\tfrac{\langle z_{x,p},z_{x,q}\rangle}{\Vert
  z_{x,p}\Vert_2\Vert z_{x,q}\Vert_2}\,\Big|\,\Vert p-q\Vert_1=\delta\Big],
  \qquad
  \mathrm{CDS}=-\beta.
  \]
  }%
  where the expectation averages over images and valid spatial pairs, $\delta$ is Manhattan
  distance on the latent grid, and $\beta$ comes from the fit $g(\delta)\simeq\alpha+\beta\delta$
  over $\delta=1,\ldots,8$ (default iREPA protocol). The cross-sectional panel uses ImageNet-val
  full50k at 256 px. Larger CDS therefore indicates faster spatial decorrelation. The figures show
  correlograms through $\delta=16$, whereas all reported CDS values use the $\delta=1,\ldots,8$
  fit.\par
\begin{figure}[H]
  \centering
  \begin{subfigure}[t]{0.48\linewidth}
  \centering
  \includegraphics[width=\linewidth]{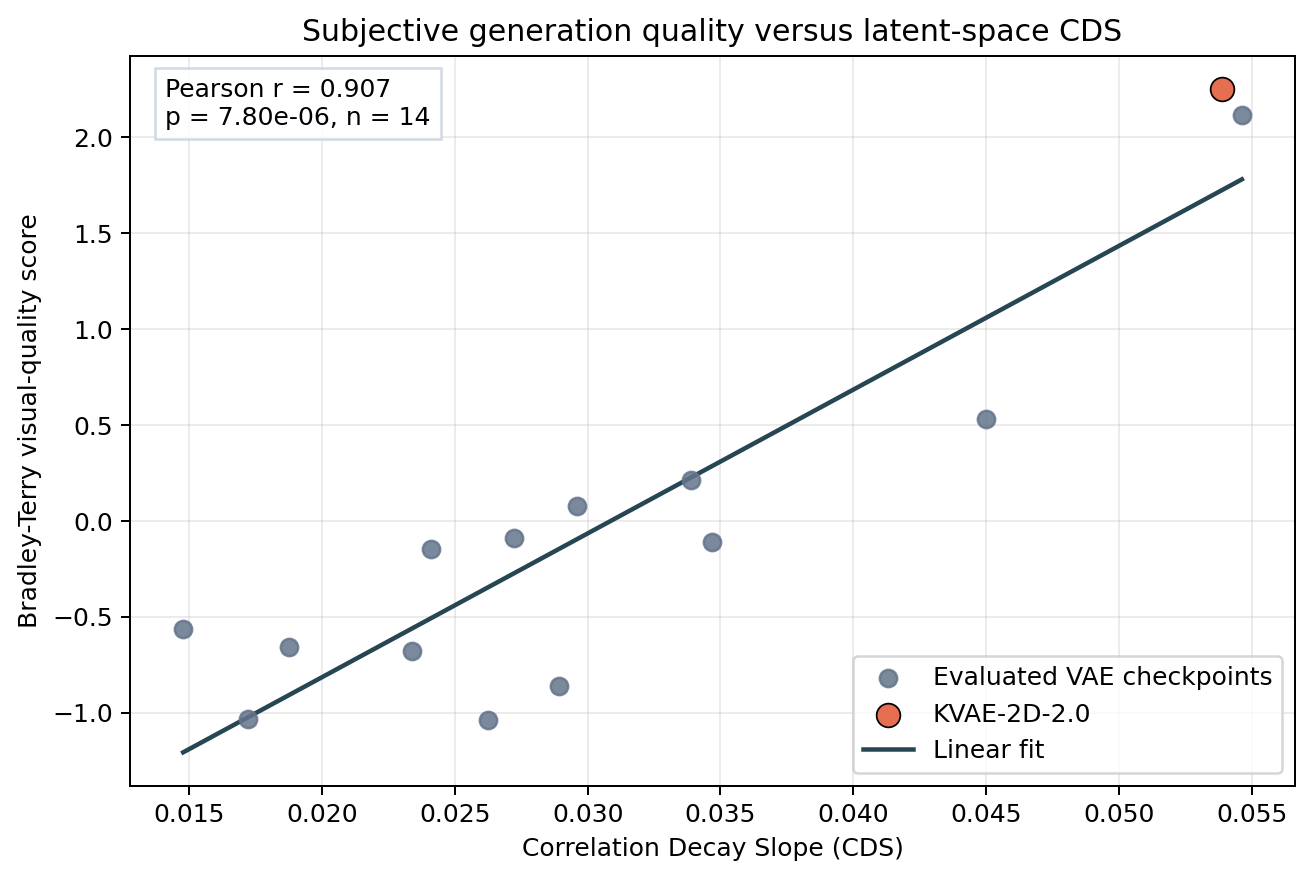}
  \caption{BT quality vs CDS.}
  \label{fig:bt_vs_cds}
  \end{subfigure}\hfill
  \begin{subfigure}[t]{0.48\linewidth}
  \centering
  \includegraphics[width=\linewidth]{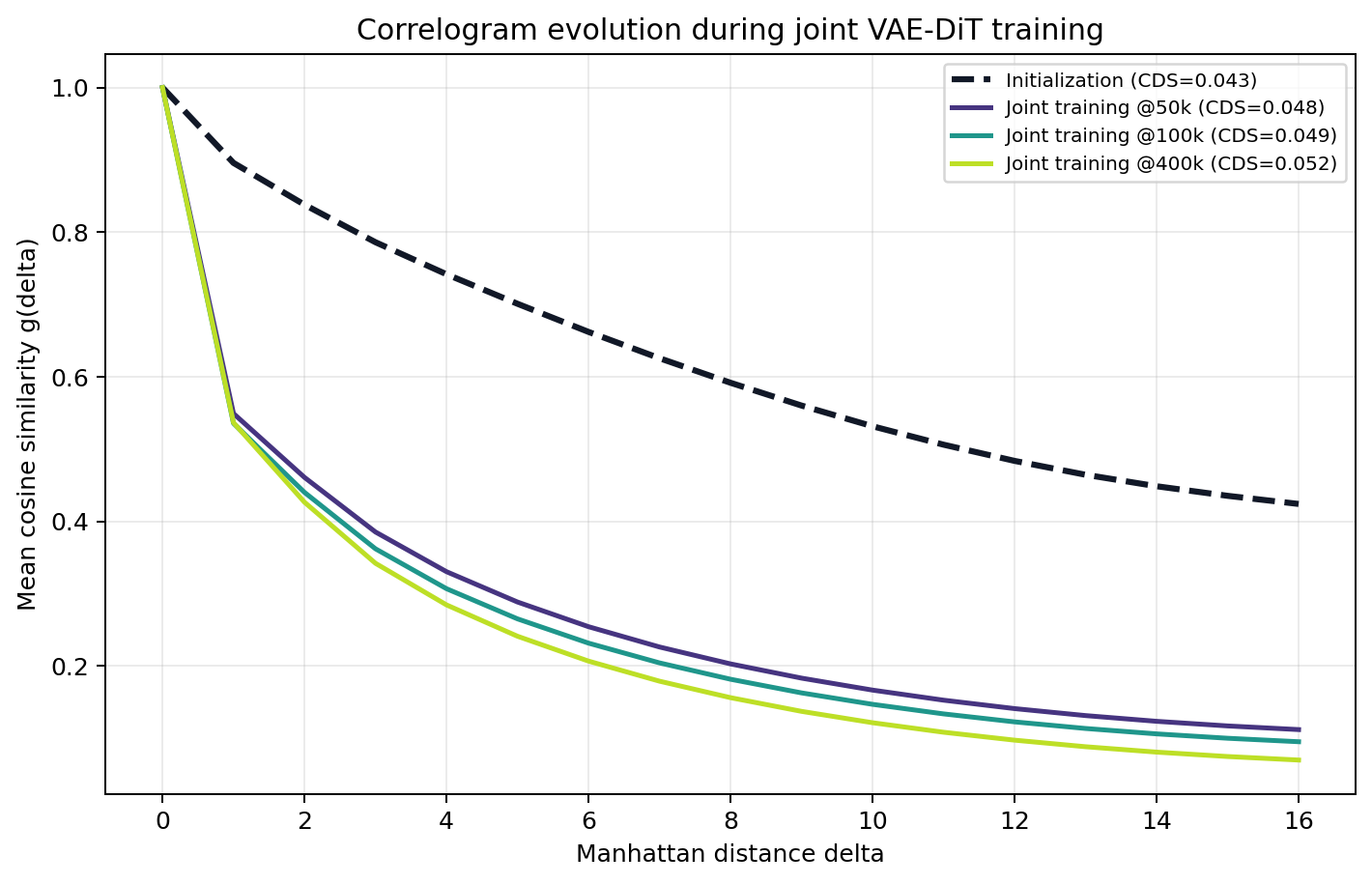}
  \caption{Joint-train correlograms.}
  \label{fig:joint_corr_evolution}
  \end{subfigure}
  \vspace{0.4em}
  \begin{subfigure}[t]{0.48\linewidth}
  \centering
  \includegraphics[width=\linewidth]{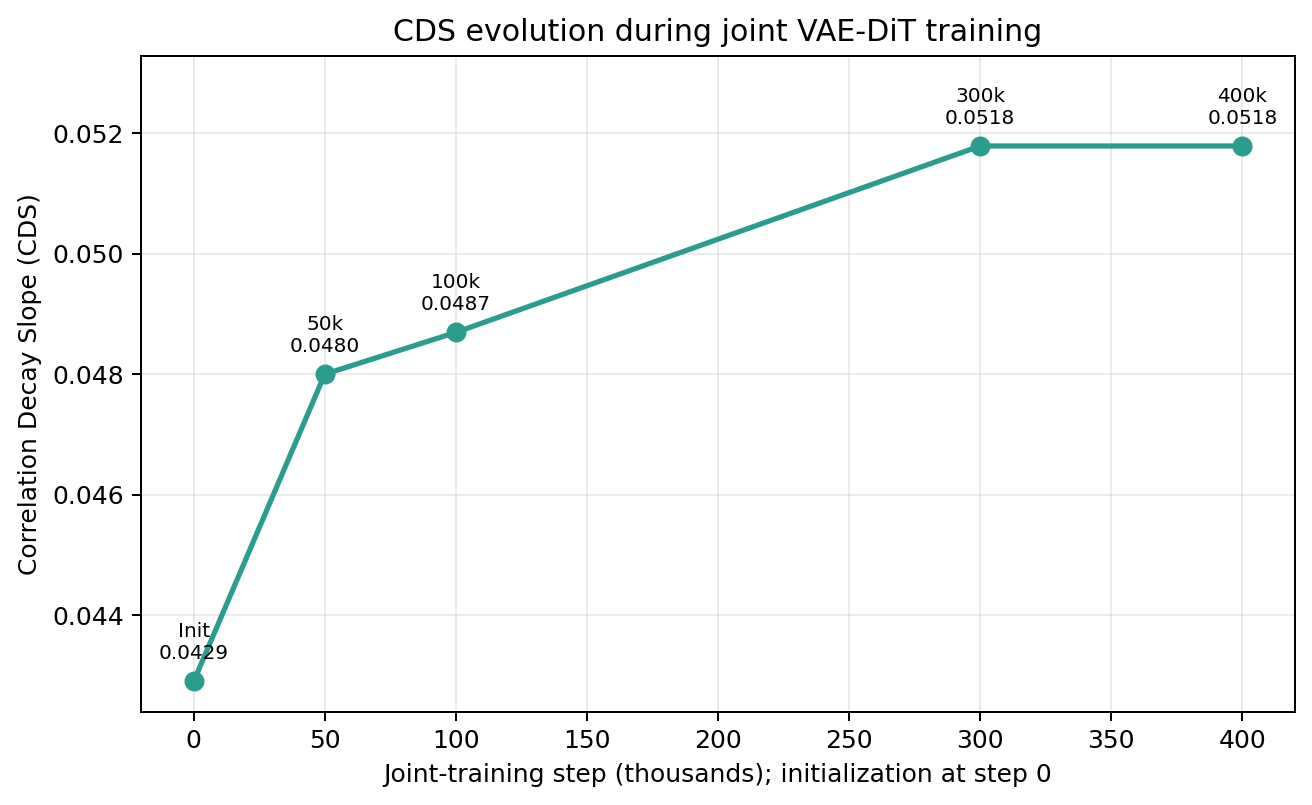}
  \caption{CDS vs joint step.}
  \label{fig:joint_cds_evolution}
  \end{subfigure}\hfill
  \begin{subfigure}[t]{0.48\linewidth}
  \centering
  \includegraphics[width=\linewidth]{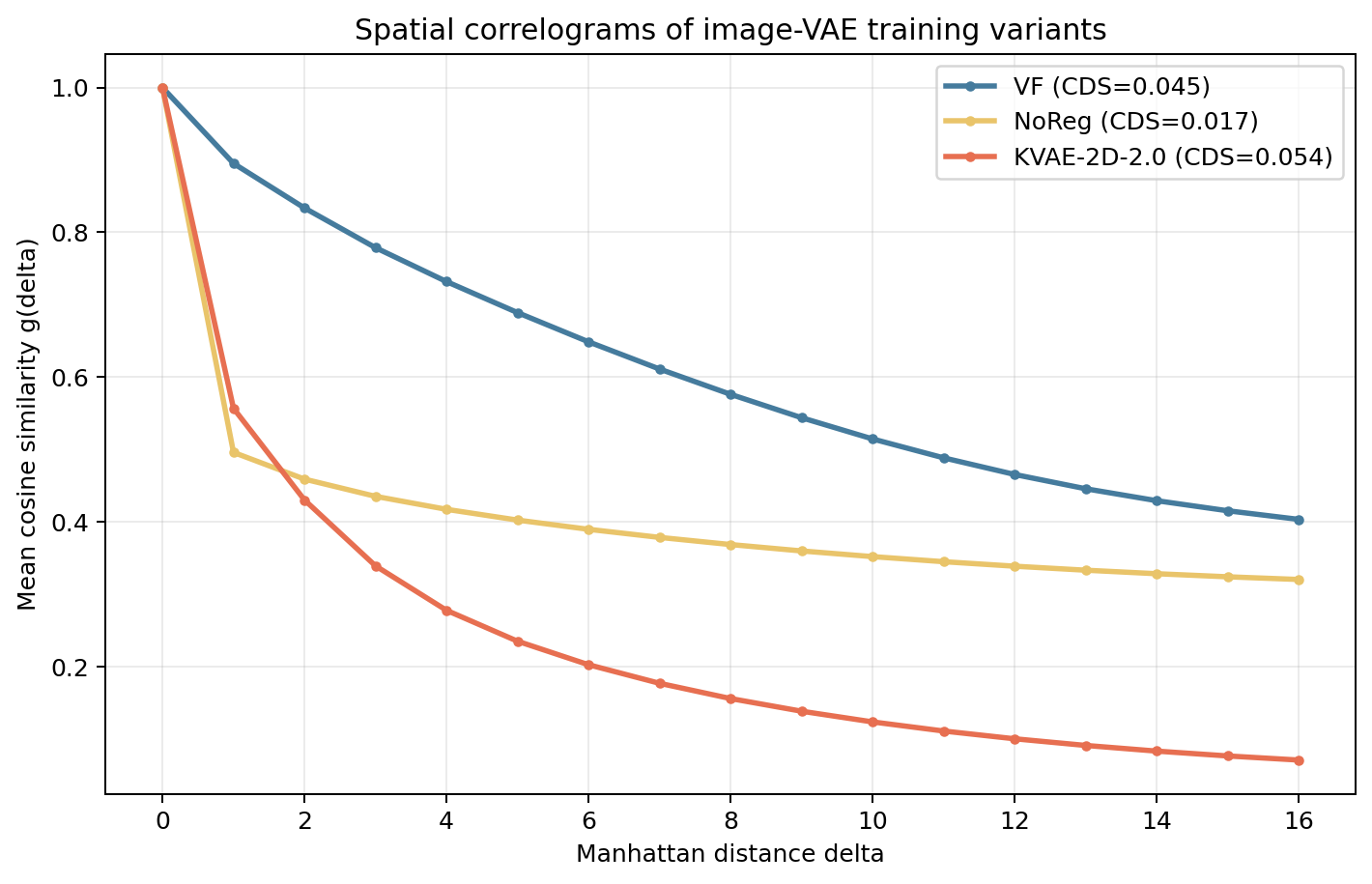}
  \caption{Tokenizer correlograms.}
  \label{fig:objective_correlograms}
  \end{subfigure}
  \caption{CDS analysis for image tokenizers: (a) Bradley--Terry visual quality vs CDS ($n{=}14$;
  KVAE-2D-2.0 highlighted); (b)--(c) correlogram and CDS along one joint VAE--DiT trajectory; (d)
  spatial correlograms for VF, NoReg (reconstruction+KL), and KVAE-2D-2.0.}
  \label{fig:cds_panel}
  \end{figure}

  CDS was introduced by iREPA for patch-token representations~\citep{singh2025irepa} and later
  applied to VAE latents in \emph{Diffusing in the Right Space}~\citep{zhong2026rightspace}.
  Across the 14 image-tokenizer configurations, CDS has Pearson correlation $r=0.906$ with the
  Bradley--Terry visual-quality score (Fig.~\ref{fig:bt_vs_cds}).  This in-sample association
  motivates its use for candidate screening, but it does not establish out-of-sample
  prediction.\par
  \myPara{CDS during joint VAE--DiT training.}
  We next examine whether CDS evolves in the direction suggested by the cross-sectional
  association.  A single trajectory optimized with a REPA-E representation-alignment
  objective~\citep{leng2025repae} is evaluated on ImageNet-val 10k using the same CDS definition.
  During this run, the correlogram becomes steeper (Figs.~\ref{fig:joint_corr_evolution}
  and~\ref{fig:joint_cds_evolution}), and CDS increases from $0.0429$ at initialization to
  $0.0518$ at 400k steps.  This one trajectory does not provide checkpoint-level validation
  against subjective quality and should not be interpreted as a general training law.\par
  \myPara{Image-tokenizer comparison.}
  Fig.~\ref{fig:objective_correlograms} compares three 32-channel 8x8 image tokenizers:
  \textbf{VF}, a vision-foundation-aligned model motivated by
  VA-VAE~\citep{yao2025reconstruction}; \textbf{NoReg}, trained with reconstruction and KL losses;
  and the final KVAE-2D-2.0 tokenizer.  The curves differ in both their absolute
  cosine-similarity levels and their rates of decay with distance.  Among these models, NoReg has
  the highest PSNR and the lowest Bradley--Terry visual-quality score, illustrating why
  reconstruction alone is insufficient for tokenizer selection.  The decay rate provides
  information not captured by a single absolute similarity value; CDS summarizes this early-range
  decay over $\delta=1,\ldots,8$.\par
  Diffusability and CDS, used as a proxy for generation performance, were crucial for training and
  selecting the best image VAEs.  The same approach was adopted for audio and video tokenizers.
  It helped identify effective designs while reducing computational cost.\par

\subsection{Design choices}
\label{sec:ablation}

\begin{figure}[H]
    \centering
    \includegraphics[width=0.95\linewidth]{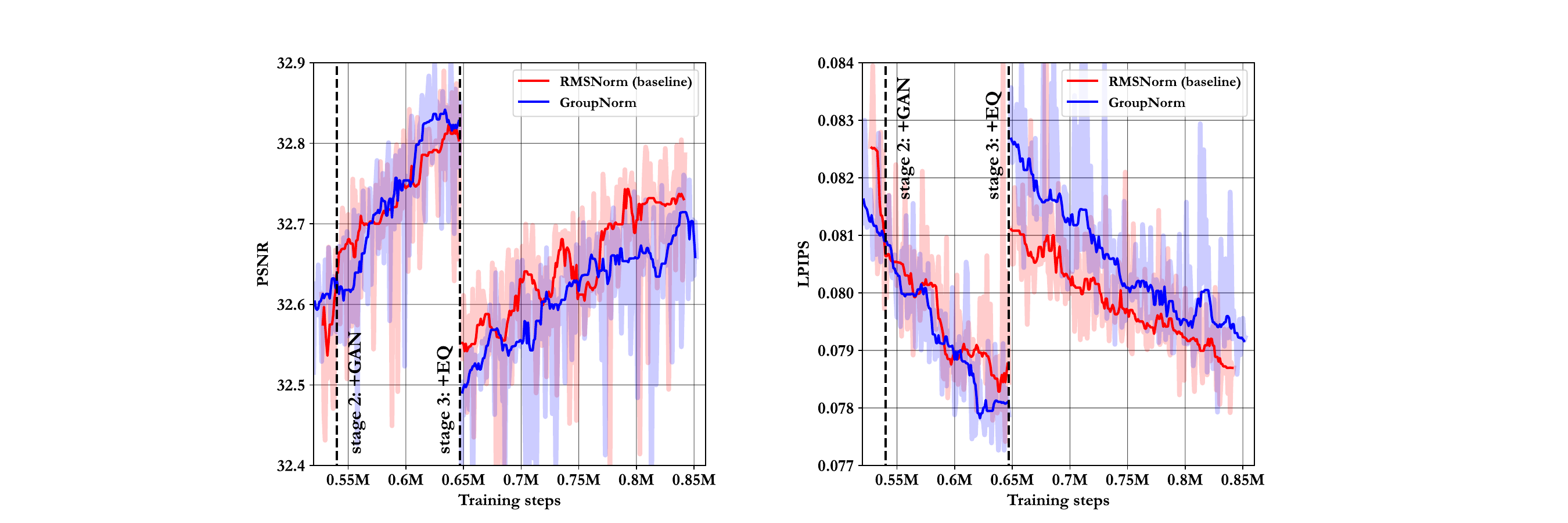}
    \caption{Normalization layer comparison on KVAE-4x8x8.}
    \label{fig:ablation_488}
\end{figure}

Original CogVideoX VAE ~\citep{cogvideox} utilized GroupNorm as normalization layer across all architecture. Since statistics were computed across all axis, including time, normalization violated causality. Moreover, during context parallel training complete computation of statistics required all-to-all synchronization. Replacing it with RMSNorm, which operate only spatially, eliminates this problems and does not increase complexity. It can be seen (Fig. \ref{fig:ablation_488}) that both models obey same learning dynamic, which allows us to switch to more practical RMSNorm without quality loss.

\begin{figure}[H]
    \centering
    \includegraphics[width=0.95\linewidth]{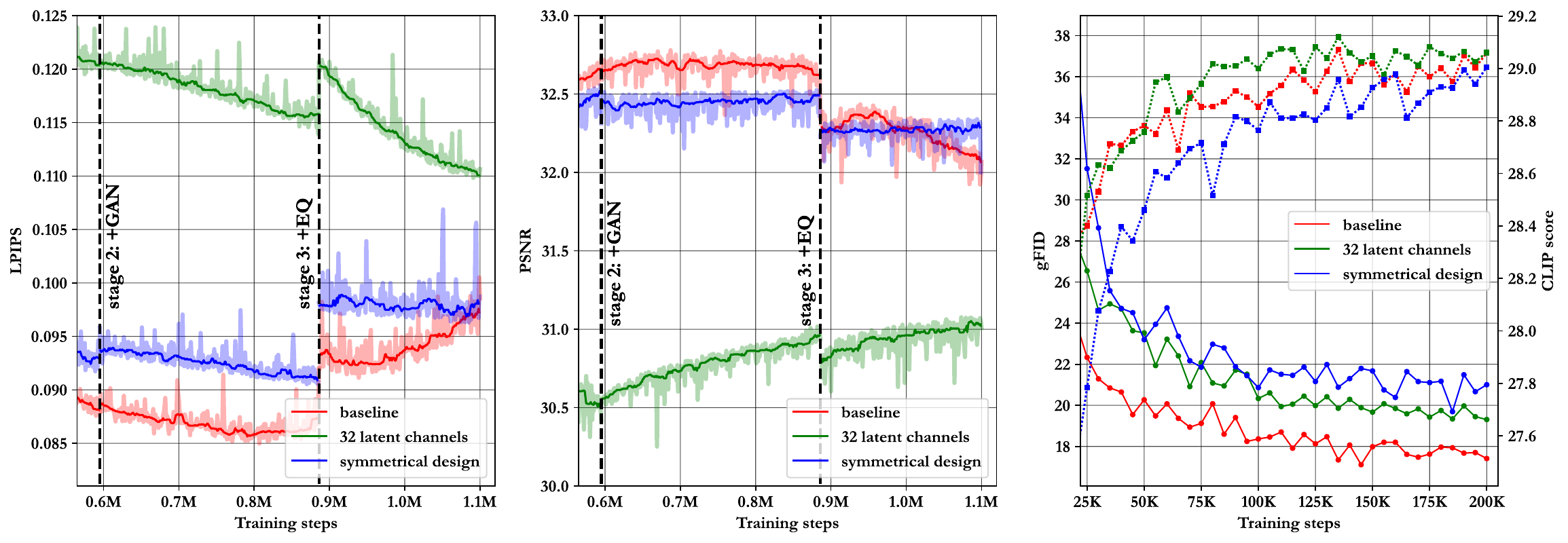}
    \caption{Ablation study for the 4x16x16 design. \textbf{Left/Middle images}: reconstruction metrics \textbf{Right image}: learning curves of text-to-video generation models}
    \label{fig:ablation_416}
\end{figure}

Width of decoder layers in KVAE-4x16x16 was increased, which affects decoding complexity and, therefore, requires justification. For that purpose version with same width in encoder and decoder was trained (see blue curve on Fig. \ref{fig:ablation_416}) and tested in image generation setting (exactly as in Sec. \ref{sec:img_gen}). Following training strategy of KVAE-4x16x16, it reaches higher reconstruction metrics, but lose pace under generation scenario, which is evident from gFID and CLIP-score plots.

While KVAE-4x16x16 is a 64 channel model, there is an argument against such large value. Main concern is amount of steps for generation model's loss to converge, and should be noted that other open models ~\citep{wan22, hunyuan15} have smaller channel size. To ablate this decision we train KVAE-4x16x16 with channels reduced down to 32 (see green curve on Fig. \ref{fig:ablation_416}). As expected, higher compression ratio leads to worse reconstruction with difference on validation up to 1dB of PSNR and 0.01 of LPIPS. But, what's more important, generation model trained with it fails to catch up with 64 channel baseline. Visual quality is consistently lags behind and CLIP-scores are close for second half of training, even reaching same values for both models on final steps.

\section{Audio tokenization}
\label{sec:audio}

Key principles of visual tokenization outlined in previous section holds for audio as well: latent diffusion is the dominant recipe for text-to-audio and joint audio-video generation, the generative model never observes a waveform, and the properties of the latent bound both what it can express and how fast it converges. What changes are the constraints. This section presents \textbf{KVAE-Audio}, a continuous full-band audio tokenizer, together with the design decisions and evaluation protocol behind it. Inference code is publicly available under the MIT license at \url{https://github.com/kandinskylab/kvae-audio}, with weights released at \url{https://huggingface.co/kandinskylab/KVAE-Audio}.

Two requirements shaped the design. The first is \textit{full-band operation}. A large share of open audio tokenizers either operates at 16--24 kHz, or compresses a mel-spectrogram and delegates waveform synthesis to a separate vocoder~\citep{audioldm2, mmaudio}. Both routes cap the usable bandwidth well below the audible range, and the latter additionally discards phase at the input and spends compute on an extra model at decoding time. For media production, where generated audio accompanies video and is mixed with speech, music and sound effects in a single track, this is unacceptable, so we model waveforms end-to-end at 48 kHz. This matters beyond audio alone: joint text-to-video-and-audio generation currently has no open solution competitive with closed ones, and a full-band tokenizer is a prerequisite rather than a refinement.

The second requirement is the \textit{diffusability} of the latent in the sense of \secref{sec:gen_perf_pred}: as in the visual domain, reconstruction quality is a poor predictor of downstream generation quality, and the two can be traded against each other. We adapted the correlation decay slope of \secref{sec:diffusability} to the one-dimensional audio latent and use it as a cheap screening signal when selecting candidates, but the final choice among them is always made after training a generative model on top of the frozen tokenizer (\secref{sec:audio_gen}). The generative model is therefore part of the evaluation protocol rather than a downstream consumer of it.

The section is structured as follows. Relevant work on audio tokenization is reviewed first (\secref{sec:audio_related}), followed by definitions, architecture and training methodology of the proposed model (\secref{sec:audio_arch} -- \secref{sec:audio_training}). Comparison with alternative solutions on reconstruction (\secref{sec:audio_rec}) and generation (\secref{sec:audio_gen}) follows, including test data and evaluation procedure, and the section closes with ablations on the design choices behind the final configuration (\secref{sec:audio_ablation}).

\subsection{Audio tokenizers for latent diffusion}
\label{sec:audio_related}

\myPara{Neural codecs.} The convolutional encoder-decoder with residual vector quantization, introduced by SoundStream~\citep{soundstream} and refined by EnCodec~\citep{encodec} and DAC~\citep{dac}, is the architectural ancestor of most modern audio tokenizers. These models are optimized for transmission: the discrete code structure and low bitrate that make a codec good make it a poor latent space for diffusion, which is why generative pipelines built on them replace quantization with a continuous bottleneck.

\myPara{Mel-spectrogram tokenizers.} An alternative line compresses mel-spectrograms rather than waveforms and restores the signal with a separate vocoder, following AudioLDM~\citep{audioldm2}. The autoencoder of MMAudio~\citep{mmaudio}, widely used in open-source text-to-audio-video pipelines, belongs to this family: a 1D convolutional VAE over mel-spectrograms producing a 40-channel latent at 43.07 Hz for 44.1 kHz input, decoded back to spectrograms and vocoded by BigVGAN-v2~\citep{bigvgan}. Same route is taken by LTX-2~\citep{ltx2} for the audio branch of a joint audio-visual model, encoding two-channel mel-spectrograms at 16 kHz into 128-dimensional tokens at 25 Hz. Phase is discarded at the input and the achievable bandwidth is bounded by the vocoder, which places these models in a different fidelity regime from end-to-end waveform tokenizers.

\myPara{Continuous waveform tokenizers.} Closest to our setting are continuous VAEs operating directly on waveforms and serving as the latent space of text-to-audio and video-to-audio models. Several of them share our starting point: MovieGen~\citep{moviegen}, HunyuanVideo-Foley~\citep{hunyuanfoley} and Qwen-Audio-VAE~\citep{qwenaudiovae} all replace the residual quantizer of DAC with a Gaussian bottleneck, arriving at 128-dimensional latents at 25 Hz (48 kHz input), 50 Hz (48 kHz) and 12.5 Hz (24 kHz) respectively, the last with a causal stack and windowed Transformers at the bottleneck for streaming use. The Stable Audio line takes a different architectural route: the Stable Audio Open VAE~\citep{stableaudioopen} is convolutional with a 64-dimensional latent at 21.5 Hz, while SAME~\citep{same}, the autoencoder of Stable Audio 3~\citep{stableaudio3}, is Transformer-based and reaches $4096\times$ compression --- roughly 10.8 Hz --- with a 256-dimensional latent over stereo 44.1 kHz input. Reported configurations therefore span 24--48 kHz in sample rate, an order of magnitude in latent frame rate, and 64--256 channels, at broadly comparable reconstruction quality; the choice of operating point is a design question rather than a matter of scale.

\myPara{Representation alignment.} Aligning latents or intermediate features to frozen foundation models has become a standard tool for improving diffusion training in the image domain~\citep{repa, yao2025reconstruction, rae}. Audio has seen analogous attempts, most of them aligning the \textit{diffusion transformer} rather than the tokenizer, and most of them targeting self-supervised speech representations~\citep{hunyuanfoley}. The closest work to ours is SAME~\citep{same}, which regularizes the autoencoder itself: semantic regression onto frozen representations, combined with a small diffusion transformer trained jointly on the latent so that its gradients shape the encoder. We explore alignment on the tokenizer side as well; see \secref{sec:audio_align}.

\subsection{Definitions and architecture}
\label{sec:audio_arch}

Define the input waveform as $x \in R^{C \times L}$, with $C$ channels of $L$ samples at a sample rate $sr$. The audio tokenizer is a pair of encoder $\mathcal{E}$ and decoder $\mathcal{D}$:
\begin{gather*}
\mathcal{E}: R^{C \times L} \rightarrow R^{d \times l} \\
\\
\mathcal{D}: R^{d \times l} \rightarrow R^{C \times L} \\
\\
l = \frac{L}{f_t}; \quad f_t = \prod_{i} s_i
\end{gather*}
where $f_t$ is the temporal compression factor, obtained as the product of layer strides $s_i$, and $d$ is the number of latent channels. Unlike the visual case, no spatial axis is present, so a tokenizer is characterized by the triple (sample rate, latent frame rate $sr/f_t$, channels $d$). KVAE-Audio operates at 48 kHz with a 50 Hz latent of 64 channels, i.e. $f_t = 960$ and one latent frame per 20 ms of audio.

KVAE-Audio inherits its convolutional backbone from DAC~\citep{dac}: residual units with dilated convolutions followed by strided convolutions that compress the signal in time and widen it in channels, with a mirrored decoder built on transposed convolutions. We keep the $\text{Snake}$ activation $x + \sin^2(\alpha x)/\alpha$, whose periodic nature is the right inductive bias for oscillatory signals and suppresses tonal artifacts. Relative to the original codec we introduce the following changes.

\myPara{Continuous bottleneck.} The residual vector quantizer and its codebooks are replaced by a reparameterization module predicting the parameters of a Gaussian posterior and sampling a continuous vector from it. This is the change that converts the codec into a VAE, and it is what makes the model usable as a tokenizer for latent diffusion at all: diffusion in a continuous space does not benefit from the discrete code structure a transmission codec is optimized for.

\myPara{Stride ladder and depth.} One block is added to both encoder and decoder, and the stride schedule is re-factorized to $[2, 3, 4, 5, 8]$ instead of the original $[2, 4, 8, 8]$, with additional projections at the encoder output and the decoder input. The product $960$ places the latent at exactly 50 Hz for a 48 kHz input, against the roughly 86 Hz of the original codec. Compared to a shorter ladder with larger individual strides, the additional stage buys compression without forcing single blocks to cover a receptive field their residual units cannot support.

\myPara{Latent channels.} The latent carries 64 channels, against 1024 in the codec it derives from and 40--256 in the continuous tokenizers we compare against. The channel count is the sharpest trade-off in the whole design and is ablated in \secref{sec:audio_ablation}.

\myPara{Attention in the bottleneck.} A single self-attention block operates on the 50 Hz sequence before the latent projection. At this frame rate attention is inexpensive --- one second of audio is 50 tokens --- and it lets the latent rely on context well beyond the local convolutional window.

\subsection{Latent representation alignment}
\label{sec:audio_align}

A recurring theme across all modalities in this report is that a tokenizer trained purely for reconstruction does not necessarily produce a latent space that is easy for a generative model to learn, and that aligning representations to a frozen foundation model is one way to improve it (\secref{sec:diffusability}). For KVAE-Audio we regularize the encoder output towards the representation of a frozen audio foundation model, having searched over the choice of that model, the layer the target is taken from, and the mechanism bringing the two representations into correspondence. The configuration we settled on improves diffusability, and with it prompt following, without a noticeable cost in reconstruction quality.

\subsection{Training methodology}
\label{sec:audio_training}

\myPara{Objectives.} Training combines a waveform-domain reconstruction term, a multi-scale spectral term, KL regularization of the posterior, an adversarial term, an auxiliary perceptual term computed on frozen audio features, and the alignment regularizer of \secref{sec:audio_align}:
\begin{equation*}
\begin{split}
L = \; & \mathcal{L}_1(x, \hat{x}) + w_{\text{spec}} \cdot \mathcal{L}_{\text{spec}}(x, \hat{x}) + w_{\text{KL}} \cdot D_{KL}(\mathcal{E}(x), \mathcal{N}(0, I)) \\
       & + w_{\text{adv}} \cdot \text{GAN}(\hat{x}, x) + w_{\text{perc}} \cdot \mathcal{L}_{\text{perc}}(x, \hat{x}) + w_{\text{align}} \cdot \mathcal{L}_{\text{align}}(\mathcal{E}(x), \mathcal{F}(x))
\end{split}
\end{equation*}
where $\hat{x} = \mathcal{D}(\mathcal{E}(x)_s)$ and $\mathcal{L}_{\text{spec}}$ is a sum of L1 distances between log-magnitude mel-spectrograms computed at several window sizes~\citep{dac}. In the last term $\mathcal{F}$ denotes a frozen audio foundation model providing the target representation and $\mathcal{L}_{\text{align}}$ the correspondence between it and the latent; the identity of $\mathcal{F}$ and the form of $\mathcal{L}_{\text{align}}$ are deferred to the dedicated publication, as stated in \secref{sec:audio_align}. Both $w_{\text{perc}}$ and $w_{\text{align}}$ are set so that neither dominates the spectral term, which we found to be the fastest way to lose high-frequency detail: the two act on the same signal from different directions, and the useful range for their sum is narrower than for either alone. All objectives are optimized by Adam~\citep{kigmaadam} with learning rates fixed within each stage.

\myPara{Discriminators.} Adversarial supervision is provided by an ensemble of discriminators of the kind established in the neural codec and vocoder literature --- period-based discriminators operating on reshaped waveform segments~\citep{hifigan} together with spectral discriminators at several STFT resolutions~\citep{dac}. We compared several such configurations, varying both the composition of the ensemble and the resolutions it covers. The comparison confirmed the general finding of that literature: no single discriminator family is sufficient on its own, since each is sensitive to a different class of artifact, and combinations covering both the temporal and the spectral view behave more stably in the late stages of training, when the remaining errors are concentrated in the upper band.

\myPara{Perceptual objectives.} For $\mathcal{L}_{\text{perc}}$ we compared several perceptual losses computed on frozen audio representations, differing in the backbone providing the features, the depth at which they are taken, and how the distance between them is formed. Perceptual terms of this kind interact strongly with the rest of the objective: they are effective at removing the artifacts that spectral distances tolerate, but a weight high enough to change the character of the reconstruction also pulls the model away from faithful high-frequency detail, so the useful range is narrow and has to be found empirically for a given backbone --- and, as noted above, jointly with the alignment weight rather than in isolation. The configuration used for the released checkpoints follows from this comparison together with the alignment study of \secref{sec:audio_align}, and is left for the dedicated publication.

\myPara{Crop length schedule.} Following the sequence-length scaling used for the video tokenizers (\secref{sec:method}), early stages train on 0.38-second crops and later stages increase the crop length to 5 seconds. Short crops maximize the number of independent examples per batch and carry the model cheaply through the bulk of convergence; long crops matter for two reasons specific to audio. They expose the model to slow envelopes --- reverberation tails, sustained tones, musical phrase structure --- that do not fit into a short window at all, and they bring the training distribution closer to the segment lengths the tokenizer will see when serving a generative model.

\myPara{Decoder fine-tuning.} The final stage freezes the encoder and fine-tunes the decoder alone. This is close to a free lunch: the latent is fixed at that point, so reconstruction improves without invalidating a generative model already trained on that latent space, and the tokenizer can keep improving after the generative training has started.

\myPara{Data.} Training data combines open and proprietary sources spanning speech, music and general sound events. One filter is worth stating explicitly: a substantial fraction of nominally 48 kHz audio in the wild is upsampled from lower rates and carries no energy in the upper band. Training a full-band tokenizer on such material teaches it to reproduce an empty spectrum, so we estimate the true bandwidth of every file and filter accordingly. After filtering, the training set amounts to roughly 10 thousand hours.

\subsection{Reconstruction}
\label{sec:audio_rec}

There is no commonly accepted benchmark for audio tokenizers, and, unlike in the visual domain, the relevant content spans domains with incompatible characteristics --- a model that reconstructs speech well may fail on dense polyphonic music or on broadband transients. We therefore evaluate on three public test sets, one per domain: AudioSet eval~\citep{audioset} for general sound, MUSDB18-HQ~\citep{musdb18hq} for music, and EARS~\citep{ears} for speech, the last being natively 48 kHz with many speakers and expressive styles.

Reported metrics cover three aspects. Spectral distances --- STFT and mel, each averaged over two window configurations and computed on linear and logarithmic magnitudes --- measure the perceptually relevant error. A sample-wise waveform distance is reported alongside them because it is the only one of the three that is sensitive to phase, which the spectral distances ignore by construction and which a waveform tokenizer is meant to preserve. Signal-level ratios (SI-SDR, SDR, SNR) quantify distortion energy, with SI-SDR invariant to overall level. For speech we additionally report PESQ-WB~\citep{pesq}.

In all reconstruction experiments the encoder output is taken as the posterior mean rather than a sample from it, for our model and for every baseline alike. Sampling would add a stochastic component that reflects the strength of the KL regularization more than the quality of the representation, and it is the mean that a generative model is trained to predict.

Baselines are open tokenizers with released weights, evaluated by us under an identical protocol: the 44.1 kHz MMAudio VAE~\citep{mmaudio}, chosen as the primary baseline because it is the autoencoder most often found under open text-to-audio-video systems; the DAC-VAE of MovieGen Audio~\citep{moviegen}, chosen because, like ours, it derives from DAC; and SAME-L~\citep{same}, the autoencoder of Stable Audio 3~\citep{stableaudio3}. Since the MMAudio VAE and SAME-L operate at 44.1 kHz, all reconstructions --- ours and MovieGen's included --- are resampled to 44.1 kHz before the metrics are computed to make the comparison fair.

\tabref{tab:audio_rec} reports the outcome. KVAE-Audio leads on spectral distances on general audio and music and is within noise of the best model on speech, while using 64 latent channels against 128 and 256 for the two waveform baselines, and 166.9M parameters against 852.1M for SAME-L. The MMAudio VAE, being mel-based, shows the expected signature of a model that does not preserve phase: competitive spectral distances alongside negative SI-SDR.

\begin{table}[htbp]
\centering
\caption{Reconstruction results by domain, all metrics computed at 44.1 kHz. Best in \textbf{bold}, second \underline{underlined}.}
\label{tab:audio_rec}
\begin{subtable}[t]{\textwidth}
\centering
\caption{AudioSet eval}
\resizebox{0.85\linewidth}{!}{%
\begin{tabular}{lcccccccc}
\toprule
\textbf{Model} & \textbf{\# Params} & \textbf{Latent dim} & \textbf{MEL$\downarrow$} & \textbf{STFT$\downarrow$} & \textbf{Waveform$\downarrow$} & \textbf{SI-SDR$\uparrow$} & \textbf{SDR$\uparrow$} & \textbf{SNR$\uparrow$} \\
\midrule
MMAudio 44.1kHz & 427.6M & 40 & \underline{0.636} & \underline{1.938} & 0.106 & -32.080 & -2.682 & -2.686 \\
DACVAE MovieGen & 107.7M & 128 & 0.669 & 2.275 & 0.029 & 8.384 & 9.421 & 9.416 \\
SAME-L & 852.1M & 256 & 0.986 & 2.726 & \underline{0.027} & \textbf{9.586} & \textbf{10.347} & \textbf{10.339} \\
KVAE-Audio & 166.9M & 64 & \textbf{0.537} & \textbf{1.770} & \textbf{0.027} & \underline{9.065} & \underline{9.920} & \underline{9.933} \\
\bottomrule
\end{tabular}%
}
\end{subtable}

\vspace{1em}

\begin{subtable}[t]{\textwidth}
\centering
\caption{MUSDB18-HQ}
\resizebox{0.85\linewidth}{!}{%
\begin{tabular}{lcccccccc}
\toprule
\textbf{Model} & \textbf{\# Params} & \textbf{Latent dim} & \textbf{MEL$\downarrow$} & \textbf{STFT$\downarrow$} & \textbf{Waveform$\downarrow$} & \textbf{SI-SDR$\uparrow$} & \textbf{SDR$\uparrow$} & \textbf{SNR$\uparrow$} \\
\midrule
MMAudio 44.1kHz & 427.6M & 40 & 0.681 & 1.865 & 0.114 & -40.204 & -3.274 & -3.273 \\
DACVAE MovieGen & 107.7M & 128 & \underline{0.519} & \underline{1.762} & 0.024 & 9.688 & 10.046 & 10.047 \\
SAME-L & 852.1M & 256 & 0.668 & 1.786 & \underline{0.023} & \underline{10.278} & \underline{10.648} & \underline{10.648} \\
KVAE-Audio & 166.9M & 64 & \textbf{0.516} & \textbf{1.725} & \textbf{0.022} & \textbf{10.390} & \textbf{10.675} & \textbf{10.677} \\
\bottomrule
\end{tabular}%
}
\end{subtable}

\vspace{1em}

\begin{subtable}[t]{\textwidth}
\centering
\caption{EARS}
\resizebox{0.95\linewidth}{!}{%
\begin{tabular}{lccccccccc}
\toprule
\textbf{Model} & \textbf{\# Params} & \textbf{Latent dim} & \textbf{MEL$\downarrow$} & \textbf{STFT$\downarrow$} & \textbf{Waveform$\downarrow$} & \textbf{SI-SDR$\uparrow$} & \textbf{SDR$\uparrow$} & \textbf{SNR$\uparrow$} & \textbf{PESQ$\uparrow$} \\
\midrule
MMAudio 44.1kHz & 427.6M & 40 & 0.616 & 1.395 & 0.030 & -29.947 & -2.728 & -2.697 & 2.424 \\
DACVAE MovieGen & 107.7M & 128 & \textbf{0.453} & \textbf{1.310} & \underline{0.006} & \textbf{10.264} & \textbf{10.680} & \textbf{10.681} & \underline{4.246} \\
SAME-L & 852.1M & 256 & 0.774 & 1.575 & 0.007 & 9.939 & 10.374 & 10.376 & 2.982 \\
KVAE-Audio & 166.9M & 64 & \underline{0.463} & \underline{1.314} & \textbf{0.006} & \underline{9.952} & \underline{10.377} & \underline{10.384} & \textbf{4.266} \\
\bottomrule
\end{tabular}%
}
\end{subtable}
\end{table}

\subsection{Generation}
\label{sec:audio_gen}

Reconstruction metrics say little about whether a latent space is a good substrate for diffusion, and, as \secref{sec:img_gen} shows for the visual case, the two can point in opposite directions. To measure the property directly we use a \textit{controlled tokenizer swap}: a text-to-audio generator is trained once per tokenizer, with the same architecture, data and number of optimizer steps, changing only the frozen autoencoder it operates on. The generator is the DiT backbone of Kandinsky-5~\citep{kandinsky5} --- latent diffusion trained with flow matching, with cross-attention to text embeddings --- re-adapted to a one-dimensional audio latent, at 0.6B parameters. Everything downstream of the tokenizer is fixed by construction, so differences in generation quality are attributable to the latent space.

Evaluation covers the same three domains as reconstruction, on AudioCaps~\citep{audiocaps} for general sound, Song Describer~\citep{songdescriber} for music, and LibriSpeech test-clean~\citep{librispeech} for zero-shot speech synthesis. We report Fr\'echet Audio Distance~\citep{fad} on three embedding backbones --- PANNs~\citep{panns}, PaSST~\citep{passt} and VGGish~\citep{vggish} --- since the metric is known to depend on the backbone and agreement across three of them guards against artifacts of any single one. CLAP score~\citep{clap} measures prompt adherence directly; on speech we additionally report WER and CER as a measure of intelligibility. CE (Content Enjoyment) and PQ (Production Quality) are the two components of Audiobox Aesthetics~\citep{audioboxaesthetics}, a predictor trained on human ratings.

Two caveats apply to \tabref{tab:audio_gen} and are worth stating rather than leaving to the reader. All three FAD backbones operate at 16--32 kHz internally, so these numbers are structurally insensitive to the band above 16 kHz: they measure whether the latent is a good substrate for generation, not whether the result is full-band. And model-based aesthetic predictors are useful as a summary of human preference but are not independent of the modelling choices of the systems they score. The side-by-side evaluation below therefore carries the main weight of the comparison rather than supplementing it.

KVAE-Audio consistently outperforms compared opensource tokenizers on AudioCaps, which is the general-domain benchmark closest to the intended use of the model. On music it leads on the aesthetic components while the mel-based MMAudio VAE retains an advantage on FAD and CLAP; on speech it leads on FAD (PANNs), on CE and on intelligibility.

\begin{table}[htbp]
\centering
\caption{Generation results under the fixed-generator tokenizer swap. Best in \textbf{bold}, second \underline{underlined}.}
\label{tab:audio_gen}
\begin{subtable}[t]{\textwidth}
\centering
\caption{AudioCaps}
\resizebox{0.9\linewidth}{!}{%
\begin{tabular}{lcccccccc}
\toprule
\textbf{Model} & \textbf{\# Params} & \textbf{Latent dim} & \textbf{CLAP$\uparrow$} & \textbf{CE$\uparrow$} & \textbf{PQ$\uparrow$} & \textbf{FAD (PANNs)$\downarrow$} & \textbf{FAD (PaSST)$\downarrow$} & \textbf{FAD (VGGish)$\downarrow$} \\
\midrule
MMAudio 44.1kHz & 427.6M & 40 & \underline{0.336} & \underline{3.909} & \underline{6.192} & \underline{17.873} & \underline{195.910} & 1.364 \\
DACVAE MovieGen & 107.7M & 128 & 0.313 & 3.772 & 6.167 & 20.558 & 234.312 & 1.700 \\
SAME-L & 852.1M & 256 & 0.322 & 3.588 & 5.756 & 18.446 & 240.635 & \underline{1.325} \\
KVAE-Audio & 166.9M & 64 & \textbf{0.344} & \textbf{3.982} & \textbf{6.242} & \textbf{15.381} & \textbf{193.760} & \textbf{1.210} \\
\bottomrule
\end{tabular}%
}
\end{subtable}

\vspace{1em}

\begin{subtable}[t]{\textwidth}
\centering
\caption{Song Describer}
\resizebox{0.9\linewidth}{!}{%
\begin{tabular}{lcccccccc}
\toprule
\textbf{Model} & \textbf{\# Params} & \textbf{Latent dim} & \textbf{CLAP$\uparrow$} & \textbf{CE$\uparrow$} & \textbf{PQ$\uparrow$} & \textbf{FAD (PANNs)$\downarrow$} & \textbf{FAD (PaSST)$\downarrow$} & \textbf{FAD (VGGish)$\downarrow$} \\
\midrule
MMAudio 44.1kHz & 427.6M & 40 & \textbf{0.356} & \underline{7.136} & \underline{7.707} & \textbf{5.412} & \textbf{158.599} & \textbf{0.356} \\
DACVAE MovieGen & 107.7M & 128 & 0.312 & 6.953 & 7.538 & 10.194 & 214.009 & 1.046 \\
SAME-L & 852.1M & 256 & \underline{0.345} & 7.076 & 7.465 & 8.442 & 250.668 & 0.987 \\
KVAE-Audio & 166.9M & 64 & 0.339 & \textbf{7.216} & \textbf{7.929} & \underline{7.971} & \underline{189.427} & \underline{0.599} \\
\bottomrule
\end{tabular}%
}
\end{subtable}

\vspace{1em}

\begin{subtable}[t]{\textwidth}
\centering
\caption{LibriSpeech test-clean}
\resizebox{\linewidth}{!}{%
\begin{tabular}{lcccccccccc}
\toprule
\textbf{Model} & \textbf{\# Params} & \textbf{Latent dim} & \textbf{CLAP$\uparrow$} & \textbf{CE$\uparrow$} & \textbf{PQ$\uparrow$} & \textbf{FAD (PANNs)$\downarrow$} & \textbf{FAD (PaSST)$\downarrow$} & \textbf{FAD (VGGish)$\downarrow$} & \textbf{WER$\downarrow$} & \textbf{CER$\downarrow$} \\
\midrule
MMAudio 44.1kHz & 427.6M & 40 & 0.368 & \underline{5.704} & 6.629 & 8.305 & \textbf{105.931} & \underline{2.001} & \underline{0.257} & \underline{0.593} \\
DACVAE MovieGen & 107.7M & 128 & \textbf{0.413} & 5.482 & \textbf{7.052} & \underline{5.008} & 210.478 & \textbf{1.501} & 0.911 & 1.048 \\
SAME-L & 852.1M & 256 & 0.379 & 4.617 & 5.024 & 10.257 & 301.508 & 2.721 & 0.349 & 0.629 \\
KVAE-Audio & 166.9M & 64 & \underline{0.389} & \textbf{5.906} & \underline{6.940} & \textbf{4.677} & \underline{185.609} & 2.138 & \textbf{0.244} & \textbf{0.576} \\
\bottomrule
\end{tabular}%
}
\end{subtable}
\end{table}

\begin{figure}[H]
\centering
\includegraphics[width=0.9\linewidth]{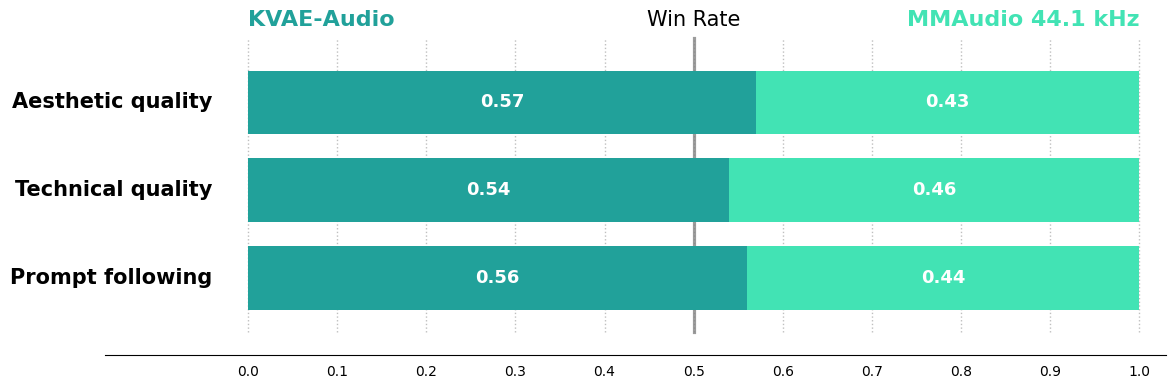}
\caption{Side-by-side evaluation of audio generations for KVAE-Audio vs MMAudio 44.1kHz}
\label{fig:audio_sbs_mmaudio}
\end{figure}

\begin{figure}[H]
\centering
\includegraphics[width=0.9\linewidth]{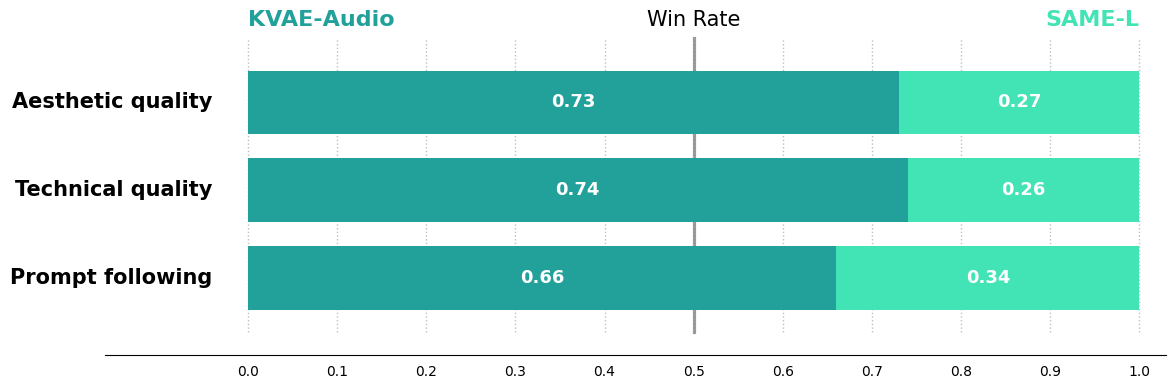}
\caption{Side-by-side evaluation of audio generations for KVAE-Audio vs SAME-L}
\label{fig:audio_sbs_same}
\end{figure}

\begin{figure}[H]
\centering
\includegraphics[width=0.9\linewidth]{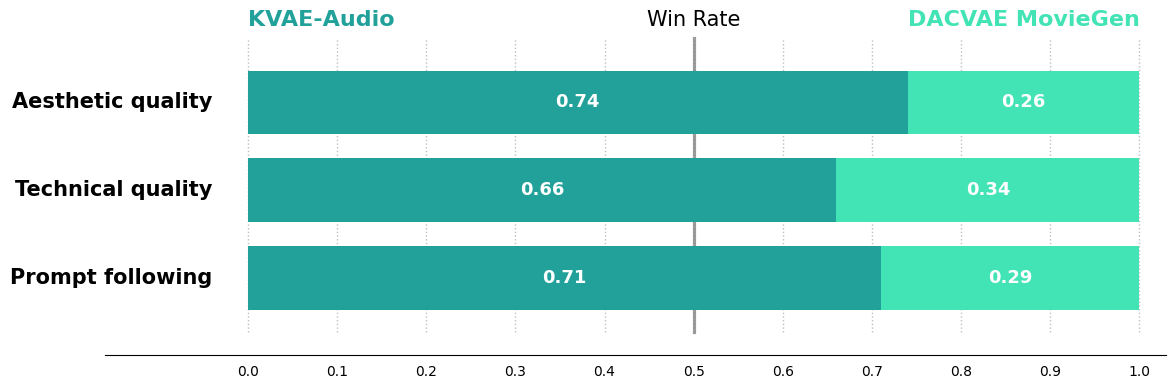}
\caption{Side-by-side evaluation of audio generations for KVAE-Audio vs DACVAE MovieGen}
\label{fig:audio_sbs_moviegen}
\end{figure}

\myPara{Side-by-side evaluation.} Following the protocol used for the visual tokenizers, annotators are shown pairs of generations produced from the same prompt by generators differing only in tokenizer, and mark a preference independently for three categories: prompt following, technical quality in the sense of freedom from noise, clipping and other artifacts, and aesthetic quality. Comparisons were run against each of the three baselines separately. Win rate is computed similar to visual tokenizers with formula from Sec. \ref{sec:img_gen}:
pairs an annotator judged equally good pull the rate toward parity, while pairs judged equally bad carry no preference signal and are dropped from the denominator. Results are given in \figref{fig:audio_sbs_mmaudio} -- \figref{fig:audio_sbs_moviegen}. KVAE-Audio is preferred over all three baselines on all three criteria, with win rates from 0.54 to 0.74. The margin is narrowest against the MMAudio VAE, the strongest baseline on the objective metrics of \tabref{tab:audio_gen}, and widest against DACVAE MovieGen and SAME-L, both of which win individual objective metrics but lose decisively to human judgement. That reversal is the reason this evaluation, rather than \tabref{tab:audio_gen}, carries the weight of the comparison.

\subsection{Design choices}
\label{sec:audio_ablation}

\myPara{Latent channel count.} The number of latent channels is the sharpest reconstruction-versus-generation trade-off we encountered. We trained variants with 16, 32, 40, 64 and 128 channels under an identical recipe and compared them through the tokenizer swap of \secref{sec:audio_gen}. Below 64 channels reconstruction quality degrades visibly; above it, reconstruction keeps improving but the diffusion model trained on the wider latent converges more slowly and reaches worse generation quality at a fixed step budget. 64 channels is the point at which the two curves cross, and it is markedly narrower than the 128 and 256 channels of the waveform baselines in \tabref{tab:audio_rec}.

One qualification on how far this number transfers. The crossing point is a property of the pair, not of the tokenizer alone: a wider latent is harder to denoise, and a generative model with more capacity has more room to absorb that difficulty, so we expect the optimal channel count to grow with the size of the model trained on top. Our generator is 0.6B parameters (\secref{sec:audio_gen}), which is modest by current standards, and 64 channels should therefore be read as the optimum at that scale rather than an intrinsic property of audio latents.

This runs opposite to the result reported for KVAE-4x16x16 in \secref{sec:ablation}, where raising the latent to 64 channels improved generation as well as reconstruction. We do not read this as a contradiction but as evidence that channel count is not the relevant quantity in isolation. The two tokenizers sit at different points on the compression-versus-information-rate curve: the video model couples its channel increase to an aggressive spatio-temporal compression factor and a token budget set by patch size, whereas the audio model already commits a comparatively high information rate per second of signal at a fixed 50 Hz frame rate, so additional channels buy reconstruction the diffusion model cannot exploit. The general form of the observation --- consistent with the reconstruction-generation dilemma~\citep{yao2025reconstruction} and with the spectral account of large bottlenecks in~\citep{improvingdiffusability} --- is that channel count must be tuned jointly with the compression factor and the resulting sequence length, and judged on generation rather than reconstruction. Reporting both modalities in one document makes the joint nature of that choice visible in a way neither result would alone.

\myPara{Attention, perceptual loss and decoder fine-tuning.} The three additions to the plain convolutional recipe were introduced cumulatively. \tabref{tab:audio_ablation} reports the relative change each one brings over a baseline trained on identical data without any of them, measured on an internal test set covering the same three domains as \secref{sec:audio_rec}; only relative changes are given, and the absolute numbers of \tabref{tab:audio_rec} remain the reproducible reference.

Attention in the bottleneck accounts for most of the spectral improvement on its own, at negligible cost at a 50 Hz frame rate --- one second of audio is 50 tokens. It does, however, couple the model's behaviour to the crop lengths seen in training, where a purely convolutional tokenizer extends to arbitrary lengths for free; in combination with the crop schedule of \secref{sec:audio_training} and windowed inference this has not been a limitation in practice, but it is the reason the decoder was left convolutional.

Adding the perceptual loss on top makes every reconstruction metric worse, and pushes the signal-level ratios marginally below the baseline. This is the trade-off described in \secref{sec:audio_training} appearing in the numbers: a term that optimizes for perceived quality is not optimizing for sample-level fidelity, and reconstruction metrics are the wrong instrument for judging it. The reason it stays in the recipe is that it improves generation quality and side-by-side preference, which \secref{sec:audio_gen} measures and this table does not.

Decoder fine-tuning then recovers the loss and goes past it, giving the best result on four of the six metrics and turning the negative signal-level ratios positive. This is the practical argument for the stage: it is applied after the latent is frozen, so it buys reconstruction quality back without touching the latent space a generative model has already been trained on.

\begin{table}[htbp]
\centering
\caption{Effect of attention in the bottleneck, the perceptual loss and decoder fine-tuning, introduced cumulatively. The baseline is the same model without all three, trained on the same data; values are relative changes with respect to it, so negative is better for the distance metrics and positive is better for the signal-level ratios. Measured on an internal test set. Best in \textbf{bold}.}
\label{tab:audio_ablation}
\resizebox{\linewidth}{!}{%
\begin{tabular}{lcccccc}
\toprule
\textbf{Configuration} & \textbf{MEL$\downarrow$} & \textbf{STFT$\downarrow$} & \textbf{Waveform$\downarrow$} & \textbf{SI-SDR$\uparrow$} & \textbf{SDR$\uparrow$} & \textbf{SNR$\uparrow$} \\
\midrule
+ attention                                          & -8.5\%          & \textbf{-8.7\%} & -3.0\%          & 2.6\%          & 1.5\%          & 1.5\%          \\
+ attention + perceptual loss                        & -5.3\%          & -6.5\%          & -1.5\%          & -0.6\%         & -0.3\%         & -0.3\%         \\
+ attention + perceptual loss + decoder fine-tuning  & \textbf{-9.5\%} & -5.2\%          & \textbf{-4.4\%} & \textbf{3.7\%} & \textbf{3.4\%} & \textbf{3.5\%} \\
\bottomrule
\end{tabular}%
}
\end{table}

\section{Conclusion}
This report presented high-quality tokenizers for rapidly developing field of text-conditioned generation.
Through series of experiments with generative models in visual and audio domains we demonstrated KVAE tokenizers surpassing leading alternatives on corresponding benchmarks.
Moreover, we've identified common patterns across modalities, exploited them in design and share our findings.
All models are released in opensource under MIT license, allowing everyone to use them in their research and incroporate into commercial solutions.

\section{Acknowledgments}

We grateful to our colleagues at Kandinsky Lab for their constant support throughout development process, especially to Denis Dimitrov as lab director, to Dmitrii Mikhailov for his work on optimization of video tokenizers which made them practical and to Vladimir Arkhipkin for insights into flow matching models and multimodal architectures.

\vspace{30pt}
{
\bibliographystyle{plain}
\bibliography{kvae,content/audio_refs}
}

\end{document}